\PassOptionsToPackage{table}{xcolor} 
\documentclass[runningheads]{llncs}
\usepackage{graphicx}
\usepackage{todonotes}
\usepackage[utf8]{inputenc} 
\usepackage{caption}
\usepackage{subcaption}
\usepackage{wrapfig}
\usepackage{booktabs} 
\usepackage{multirow} 
\usepackage{makecell} 
\usepackage{array}
\newcommand{\cfbox}[2]{%
	\colorlet{currentcolor}{.}%
	{\color{#1}%
 	    \setlength{\fboxsep}{0pt}%
		\setlength{\fboxrule}{0.5pt}%
		\fbox{\color{currentcolor}#2}}%
	}
\usepackage{dirtree}
\usepackage{fancyvrb}

\begin{document}

\rowcolors*{1}{}{gray!15} 
\newcommand{\gc}{\cellcolor{gray!15}} 

\newcommand\myicon[1]{{\color{#1}\rule{2ex}{2ex}}}
\newcommand{\myfolder}[2]{\myicon{#1}\ {#2}}

\title{GLENDA: Gynecologic Laparoscopy Endometriosis Dataset}
%
%

\author{Andreas Leibetseder\inst{1}\orcidID{0000-0002-9535-966X} \and
Sabrina Kletz\inst{1}\orcidID{0000-0002-7275-0594} \and
Klaus Schoeffmann\inst{1}\orcidID{0000-0002-9218-1704} \and
Simon Keckstein\inst{2} \and \\ J{\"o}rg Keckstein\inst{3}}
\authorrunning{A. Leibetseder et al.} 
\institute{Institute of Information Technology, Klagenfurt University,\\Klagenfurt 9020, Austria\\\email{\{aleibets,sabrina,ks\}@itec.aau.at} \and
University Hospital, Ludwig-Maximilians-University Munich,\\Munich 80799, Germany\\\email{simon.keckstein@med.uni-muenchen.de} \and
Medical Faculty, Ulm University,\\Ulm 89081, Germany\\\email{joerg@keckstein.at}}

\maketitle              
\begin{abstract}
Gynecologic laparoscopy as a type of minimally invasive surgery (MIS) is performed via a live feed of a patient's abdomen surveying the insertion and handling of various instruments for conducting treatment. Adopting this kind of surgical intervention not only facilitates a great variety of treatments, the possibility of recording said video streams is as well essential for numerous post-surgical activities, such as treatment planning, case documentation and education. Nonetheless, the process of manually analyzing surgical recordings, as it is carried out in current practice, usually proves tediously time-consuming. In order to improve upon this situation, more sophisticated computer vision as well as machine learning approaches are actively developed. Since most of such approaches heavily rely on sample data, which especially in the medical field is only sparsely available, with this work we publish the Gynecologic Laparoscopy ENdometriosis DAtaset (GLENDA) -- an image dataset containing region-based annotations of a common medical condition named endometriosis, i.e. the dislocation of uterine-like tissue. The dataset is the first of its kind and it has been created in collaboration with leading medical experts in the field.

\end{abstract}

\section{Introduction}

Minimally invasive surgery (MIS) considerably reduces trauma inflicted upon patients during medical interventions, since, as opposed to traditional open surgery, treatments are applied less intrusively. As a typical form of MIS, \textit{endoscopy} is performed by inserting a small camera, the \textit{endoscope}, as well as a variety of instruments into the human body via natural or artificially created orifices. In the case of \textit{gynecologic laparoscopy} such incisions are created into the abdomen in order to treat conditions related to the female reproductive system. The accordingly obtained video feed of an individual's inner anatomy is projected onto external monitors providing physicians with adequate visuals for performing surgery.

With the prospect of conducting surgeries in such a manner comes the possibility of recording entire procedures, an opportunity that is in fact pursued by most modern medical facilities. Apart from representing valuable evidence for lawful investigations, these kind of recordings more importantly are consulted by medical practitioners for further treatment planning, case revisitations or even educational purposes. Seemingly a convenient improvement, several downsides, however, considerably diminish the usefulness of archived surgery footage: recording the typically hours-long surgeries filmed in high-definition on a daily basis requires elaborate long-term storage solutions. Furthermore, in order to remain useful, video archives of such magnitudes must easily be searchable even by potentially non tech-savvy staff. The consequentially arising need for more sophisticated systems capable of aiding physicians post- as well as even intra-surgery creates great opportunities and challenges for various scientific communities, not least the ones concerned with multimedia~\cite{Munzer2017}.

Although machine learning has successfully been applied in the field of medical imaging~\cite{litjens2017survey}, specifically for the task of disease classification and diagnosis, much needed published datasets for feeding corresponding algorithms are only sparsely available. This not only is due to the increased sensitiveness of such data but as well a consequence of the broad spectrum of different imaging technologies\footnote{e.g. X-ray, computed tomography (CT) scans, magnetic resonance imaging (MRI), ultrasound, ...} utilized for a great variety of purposes. When merely regarding endoscopy in general the number of publically available datasets is reduced even more drastically, leaving only a few for the sub-discipline of laparoscopy: Cholec80~\cite{Twinanda2017}, LapChole~\cite{stauder2016tum}, GI dataset~\cite{ye2016online}, SurgicalActions160~\cite{Schoeffmann2018} and LapGyn4~\cite{leibetseder2018lapgyn4} to name some recent ones.

\begin{figure}[!htb]
\centering
\begin{subfigure}{.5\textwidth}
  \centering
  \includegraphics[width=.9\linewidth]{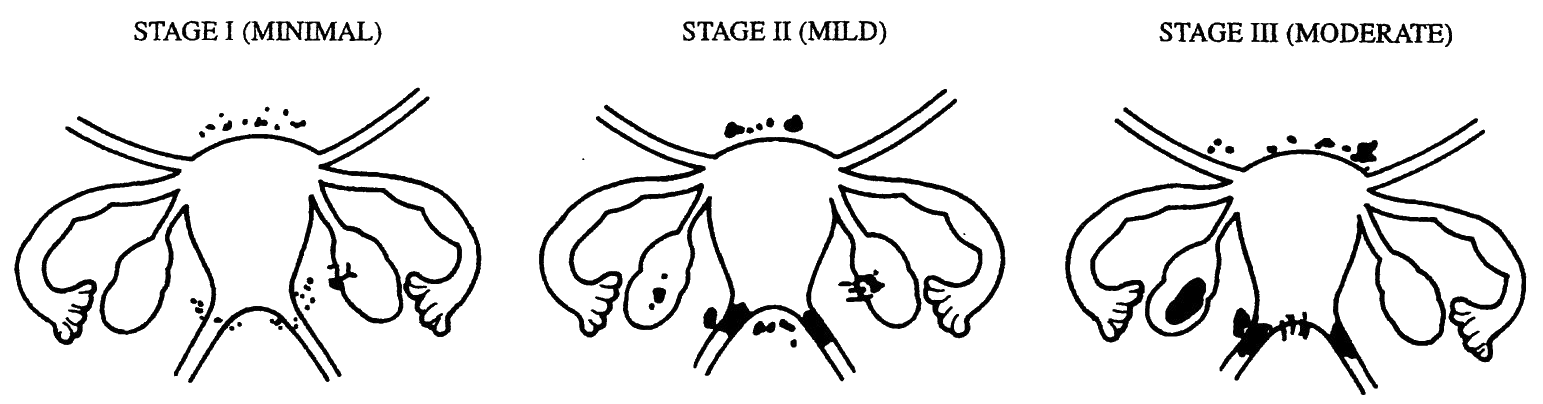}
  \caption{rASRM examples (peritoneum, ovary) of varying severity}
  \label{fig:rasrm}
\end{subfigure}%
\begin{subfigure}{.5\textwidth}
  \centering
  \includegraphics[width=.9\linewidth]{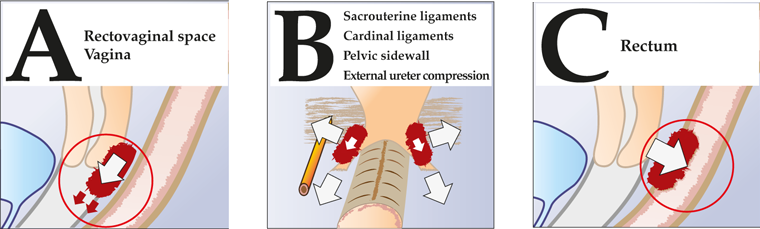}
  \caption{Enzian examples}
  \label{fig:enzian}
\end{subfigure}
\caption{Example endometriosis locations for rASRM (a) and Enzian (b).}
\label{fig:rasrm_enzian}
\vspace{-4mm}
\end{figure}

As cholecystectomy, i.e. the removal of the gallbladder, is the most frequently conducted laparoscopic surgery~\cite{tsui2013minimally}, released datasets commonly are created from this procedure. Keeping that in mind, with this work we specifically target  a different kind of procedure, which typically as well is treated laparoscopically: the diagnosis, inspection and surgical removal of \textit{endometriosis} -- a benign but painful anomaly among women in child bearing age involving the growth of uterine-like tissue in locations outside of the uterus. The condition can be found in various positions and severities, often in multiple instances per patient requiring a physician to determine its extent. This most frequently is accomplished by calculating its magnitude via utilizing the combination of two popular classification systems, the revised American Society for Reproductive Medicine (rASRM) score~\cite{canis1997revised} and the European~\cite{andrews1985revised} Enzian classification~\cite{Keckstein2017}, which describe the anomaly's potential anatomical location and severity on a three-level scale. Figure~\ref{fig:rasrm_enzian} shows a few examples of these locations as described by both of these systems. Our contribution, the Gynecologic Laparoscopy ENdometriosis DAtaset\footnote{%
\url{http://www.itec.aau.at/ftp/datasets/GLENDA}
} (GLENDA) dataset comprises a subset of these locations and has been created with leading medical experts in the field of endometriosis treatment. The dataset and with it our contribution can be characterized as follows:

\begin{description}
    \item[Source] 300+ video segments and frames selected from a pool of 400+ individual full surgery videos.
    \item[Images] 25K+ Images, consisting of 12K+ positive, i.e. pathological images associated with endometriosis, and 13K+ negative examples, i.e. non-pathological images without visible endometriosis.
    \item[Annotations] 500+ hand-drawn region-based class-specific endometriosis annotations on 300+ images/keyframes.
    \item[Classes] Five pathological categories, of which four are based on the location of the condition (peritoneum, ovary, uterus, DIE -- deep infiltrating endometriosis) and one indicating no visible endometriosis (no pathology).
    \item[Purposes] Binary as well as multi-label (endometriosis) classification, detection and localization tasks with the option of tracking pathology over video segments, thus, augmenting the overall annotated sample count. 
\end{description}

The remainder of this paper discusses details about the dataset, starting with its creation in Section~\ref{sec:dataset_creation}, its structure in Section~\ref{sec:the_dataset} and, finally, discussing its limitations in Section~\ref{sec:limitations} before drawing conclusions in Section~\ref{sec:conclusion}.

\section{Dataset Creation}
\label{sec:dataset_creation}

Overall, there are very numerous potential locations for endometriosis\footnote{e.g. peritoneum, ovary, tube, ligaments, vagina, rectum, bladder, ureter, ...} and the condition's size determines its severity level on a scale from one to three. Hence, for creating a a complete dataset in terms of a sufficient amount of examples for every possible combination of location and severity extent requires the collection of samples for well over 50 different category types or classes. This, in fact, can be considered an overly challenging task due to the following reasons:

\begin{description}
    \item[Expert Knowledge] Endometriosis can not reliably be recognized by laymen or even untrained medical practitioners, which stresses the need for employing specific experts in the field and at the same time severely reduces the amount of capable annotators for such a dataset.
    \item[Time] GLENDA has been created with fully active surgeons restricting all annotation and research effort to non-working hours, which considerably slows down the data collection process.
    \item[Rarity] Several lesion locations are diagnosed much more rarely than others, hence, including them prolongs data accumulation even further.
    \item[Completeness] Finding representative examples of a class for each of the three potential severity levels even for a reduced set of categories poses an additional challenge outweighing the time and effort for spent annotating, at least for the dataset's current first version.
\end{description}

Therefore, for our initial GLENDA version we constrain annotations to a subset of four endometriosis locations (see Section~\ref{sec:the_dataset} for details) consisting of region-based annotations of single video frames, which either are associated with specific video positions (frame annotations) or sequences over time (keyframe annotations in video segments). Although for video segments only keyframes are annotated, they have been created keeping in mind that all of the endometriosis regions identified on them are visible throughout the sequences, i.e. camera motion is kept at a minimal level. This offers the possibility of augmenting the number of annotations by applying annotation tracking mechanisms to these segments (e.g. point/kernel/silhoutte tracking).

\begin{figure}[!htb]
\centering
\begin{subfigure}{.5\textwidth}
  \centering
  \includegraphics[width=0.95\linewidth]{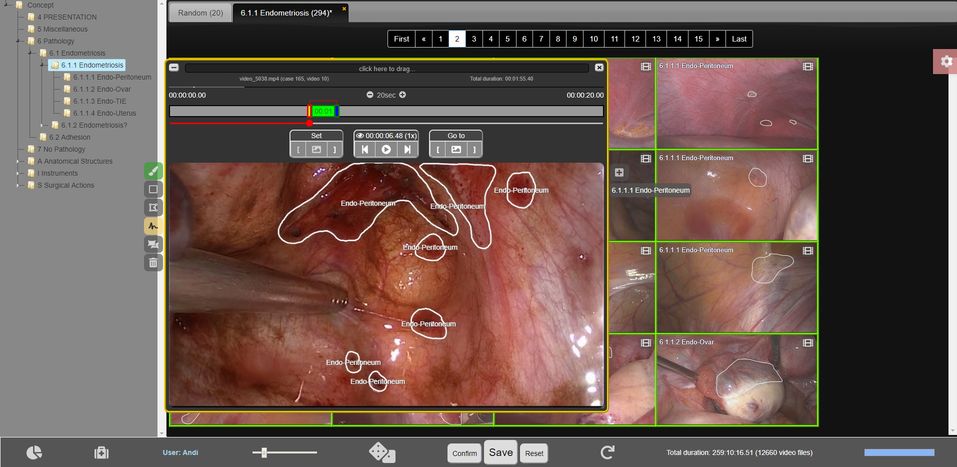}
  \caption{Creating annotations using closed free-hand drawings, polygons and rectangles.}
  \label{fig:ecat_create}
\end{subfigure}%
\begin{subfigure}{.5\textwidth}
  \centering
  \includegraphics[width=0.95\linewidth]{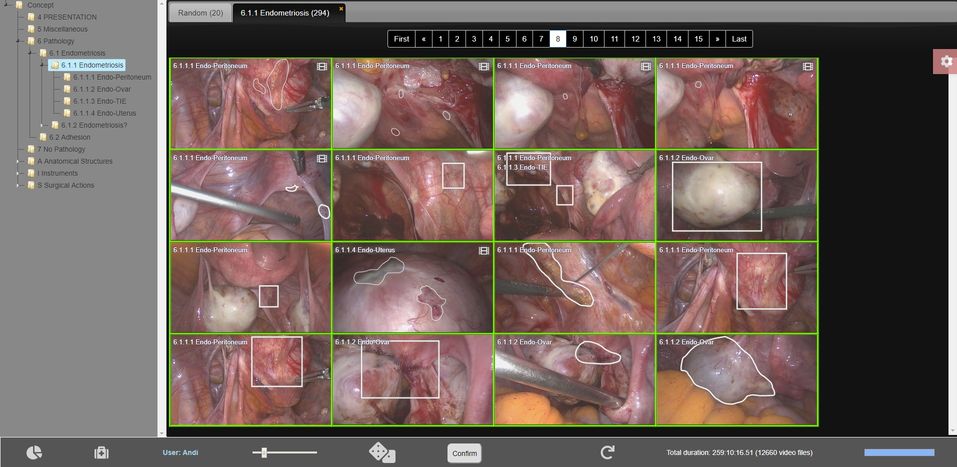}
  \caption{Different endometriosis annotations in a summary view.}
  \label{fig:ecat_explore}
\end{subfigure}
\caption{Dataset creation and exploration using the Endoscopic Concept Annotation Tool (ECAT).}
\label{fig:ecat}
\vspace{-4mm}
\end{figure}

The entire dataset has been created using the Endoscopic Concept Annotation Tool~\cite{munzer2019ecat} (ECAT), shown in Figure~\ref{fig:ecat}. ECAT is a web-technologies-based tool that allows for importing large video databases and creating concept annotations for video sequences as well as frames. It in particular enables users to annotate frames by creating rectangles as well as closed polygons and free-hand drawings, meaning that every annotation always needs to enclose a region. Further, a single annotation on a frame that potentially can contain many different annotations, is required to be associated with one of GLENDA's four endometriosis categories. For video sequences, a keyframe must be chosen first before being able to draw an annotation region. 

As the system can be utilized via a standard web browser, it has been made remotely accessible to all involved medical experts for increasing the convenience when creating annotations. Finally, all data for the current GLENDA version has been collected over the course of four months time.

\section{The GLENDA Dataset}
\label{sec:the_dataset}

GLENDA is a multi-faceted endometriosis dataset that has been extracted from over 400 gynecologic laparoscopy videos, many of which show endometriosis cases of varied severities. It is summarized in Table~\ref{tab:dataset_statistics} and composed of following elements:

\begin{description}
    \item[Categories] Five categories/classes describe a distinct endometriosis locations:
    \begin{itemize}
        \item pathology: peritoneum, ovary, uterus, deep infiltrating endometriosis (DIE)
        \item no pathology: (no visible endometriosis)
    \end{itemize}
    \item[Annotations] Region-based as well as temporal annotations in the form of:
    \begin{itemize}
        \item Annotated frames: single video frames annotated with hand-drawn sketches (regions), which indicate one or more out of four endometriosis categories.
        \item Annotated sequences: sets of consecutive video frames associated with one or several categories over certain periods of time with annotated keyframes (pathology) or no additional region-based annotations (no pathology).
    \end{itemize}
\end{description}

\setlength{\tabcolsep}{4.5pt}
\renewcommand{\arraystretch}{1.5}
\begin{table}[!htb]
\centering
\caption{GLENDA summary: number of annotations (annot.) per category (cat.), number of annotated frames per category, maximum (max.) annotations per frame, max. categories (cat.) per frame, number of sequences (seq.) and amount of frames.}
\begin{tabular}{lrccccc}
\toprule
\hiderowcolors
\multicolumn{2}{c}{\textbf{Category}\textsuperscript{*}} & \textbf{annot.} & \makecell{\textbf{annot.} \\ \textbf{frames}} & \makecell{\textbf{max. annot.} \\ \textbf{per frame}} & \textbf{seq.} & \textbf{frames}\\
\midrule
 & \gc peritoneum & \gc 402 & \gc 203 & \gc 9 & \gc 73 & \gc 6470\\
 &  ovary & 51 & 48 & 2 & 15 & 2478 \\
 & \gc uterus & \gc 14 & \gc 8 & \gc 3 & \gc 5 & \gc 475 \\
\multirow{-4}{*}{pathology} & DIE & 53 & 43 & 3 & 18 & 2821 \\
\midrule
no pathology & & 0 & 0 & 0 & 27 & 13 438 \\
\midrule
\textbf{Total} & & \textbf{520} & \textbf{302} & \textbf{9 (max. cat.: 3)} & \textbf{138} & \textbf{25 682} \\
\bottomrule
\end{tabular}
\vspace*{-2mm}
{\begin{flushleft}\scriptsize\textsuperscript{*}Note: A sequence/keyframe/frame is attributed to a specific category if it is the dominant one in all of its corresponding annotations in terms of annotation count/area covered.\end{flushleft}}
\label{tab:dataset_statistics}
\end{table}

As a consequence of choosing above structure, the dataset allows for a multitude of utilization purposes. Splitting up the dataset into pathology and no pathology images allows for attempting binary classification, disregarding all endometriosis sub-classes. Furthermore, when including individual class annotations multi-class endometriosis prediction can be approached, as already mentioned in above Section~\ref{sec:dataset_creation} potentially by augmenting the amount of annotations via tracking them throughout their corresponding video segments. Aside from possible disadvantages outlined in Section~\ref{sec:limitations}, additionally collecting video segments has the advantage of enabling the inclusion of temporal information in proposed methodologies for analysis. Finally, the multitude of region-based annotations can be leveraged for localization tasks as well as representing a basis for learning further annotations.

Following sections more thoroughly describe GLENDA's class categories as well as structure on a file basis, while pointing out the dataset's limitations.
 
\subsection{Categories}
\label{sec:ds_classes}


\begin{figure}[!htb]
	\begin{subfigure}[b]{0.18\textwidth}
		\centering
		\cfbox{black}{\includegraphics[width=\textwidth]{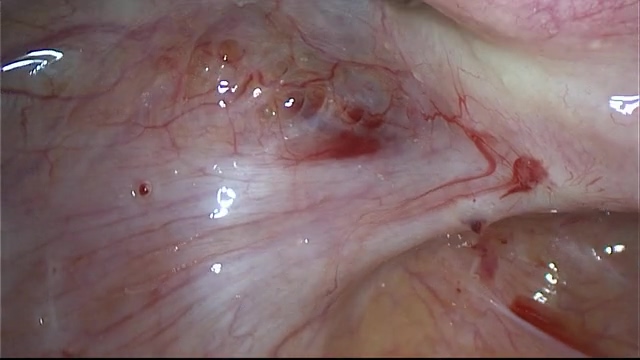}}
		\caption[]%
		{{\small}}    
		\label{fig:per_top_0}
	\end{subfigure}
	\hskip0.5em\relax
	\begin{subfigure}[b]{0.18\textwidth}   
		\centering 
		\cfbox{black}{\includegraphics[width=\textwidth]{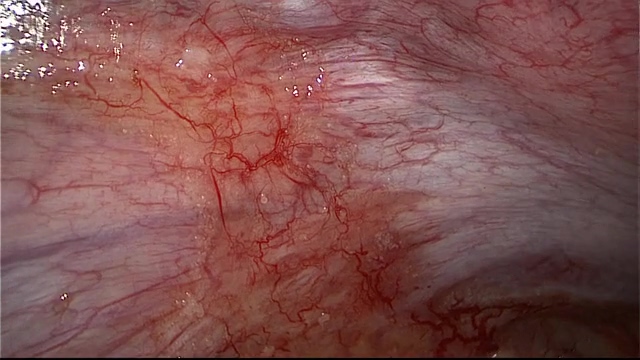}}
		\caption[]%
		{{\small}}    
		\label{fig:per_top_1}
	\end{subfigure}
	\hskip0.5em\relax
	\begin{subfigure}[b]{0.18\textwidth}   
		\centering 
		\cfbox{black}{\includegraphics[width=\textwidth]{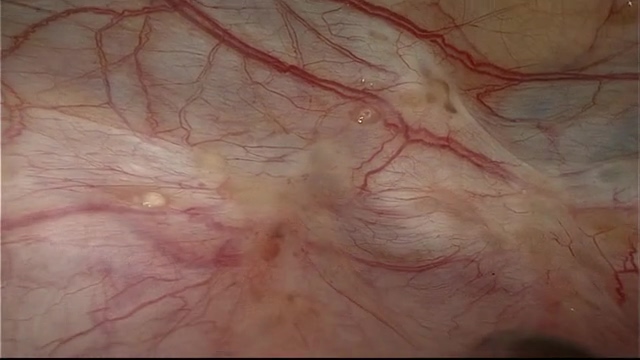}}
		\caption[]%
		{{\small}}    
		\label{fig:per_top_2}
	\end{subfigure}
	\hskip0.5em\relax
	\begin{subfigure}[b]{0.18\textwidth}
		\centering
		\cfbox{black}{\includegraphics[width=\textwidth]{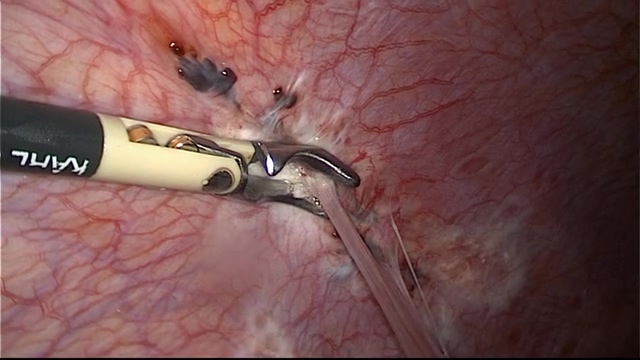}}
		\caption[]%
		{{\small}}    
		\label{fig:per_top_3}
	\end{subfigure}
	\hskip0.5em\relax
	\begin{subfigure}[b]{0.18\textwidth}   
		\centering 
		\cfbox{black}{\includegraphics[width=\textwidth]{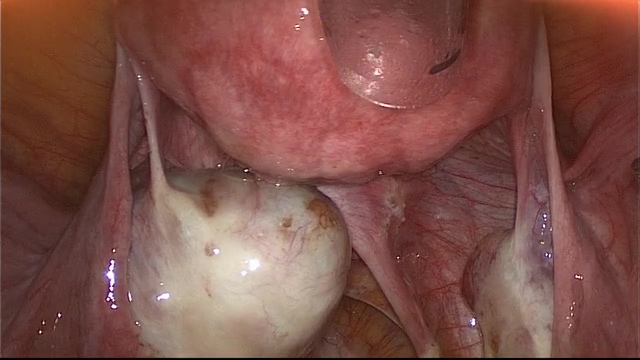}}
		\caption[]%
		{{\small}}    
		\label{fig:per_top_4}
	\end{subfigure}
	\vskip0.5em\relax
	\begin{subfigure}[b]{0.18\textwidth}
		\centering
		\cfbox{black}{\includegraphics[width=\textwidth]{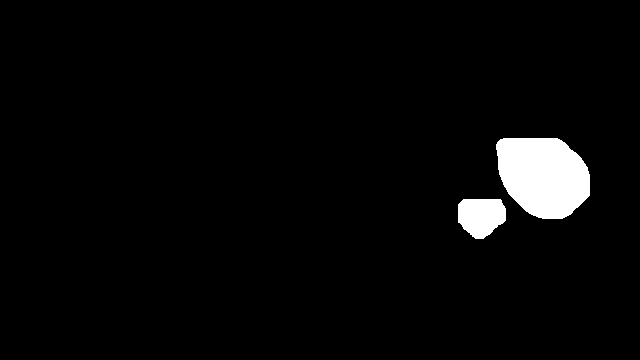}}
		\caption[]%
		{{\small}}    
		\label{fig:per_mid_0}
	\end{subfigure}
	\hskip0.5em\relax
	\begin{subfigure}[b]{0.18\textwidth}   
		\centering 
		\cfbox{black}{\includegraphics[width=\textwidth]{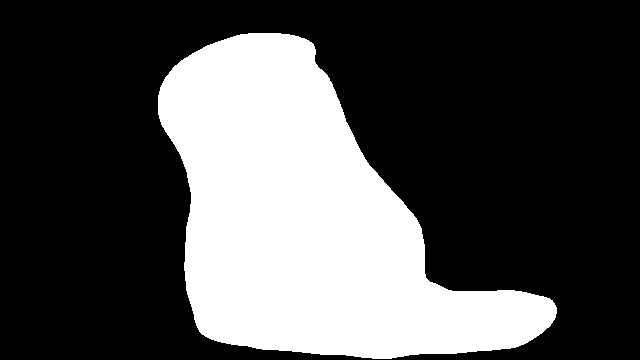}}
		\caption[]%
		{{\small}}    
		\label{fig:per_mid_1}
	\end{subfigure}
	\hskip0.5em\relax
	\begin{subfigure}[b]{0.18\textwidth}   
		\centering 
		\cfbox{black}{\includegraphics[width=\textwidth]{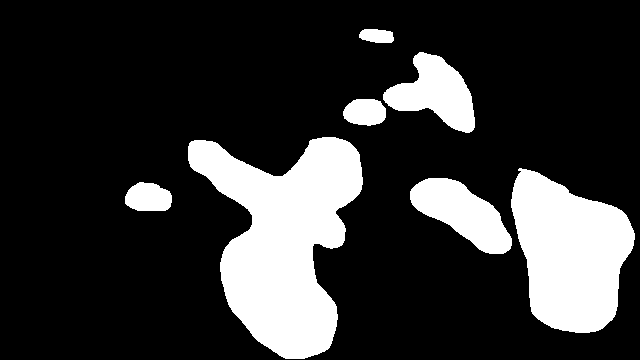}}
		\caption[]%
		{{\small}}    
		\label{fig:per_mid_2}
	\end{subfigure}
	\hskip0.5em\relax
	\begin{subfigure}[b]{0.18\textwidth}
		\centering
		\cfbox{black}{\includegraphics[width=\textwidth]{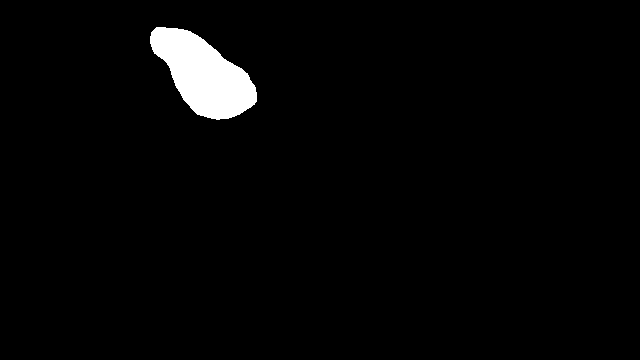}}
		\caption[]%
		{{\small}}    
		\label{fig:per_mid_3}
	\end{subfigure}
	\hskip0.5em\relax
	\begin{subfigure}[b]{0.18\textwidth}   
		\centering 
		\cfbox{black}{\includegraphics[width=\textwidth]{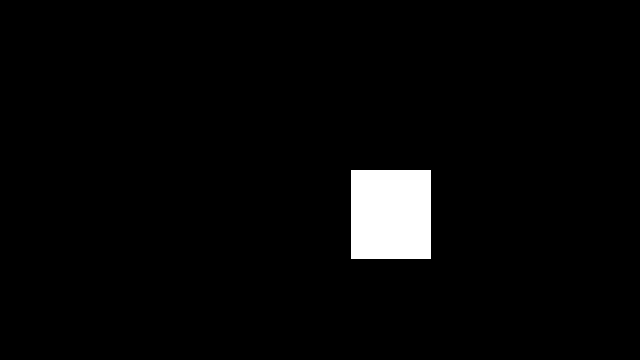}}
		\caption[]%
		{{\small}}    
		\label{fig:per_mid_4}
	\end{subfigure}
	\vskip0.5em\relax
		\begin{subfigure}[b]{0.18\textwidth}
			\centering
			\cfbox{black}{\includegraphics[width=\textwidth]{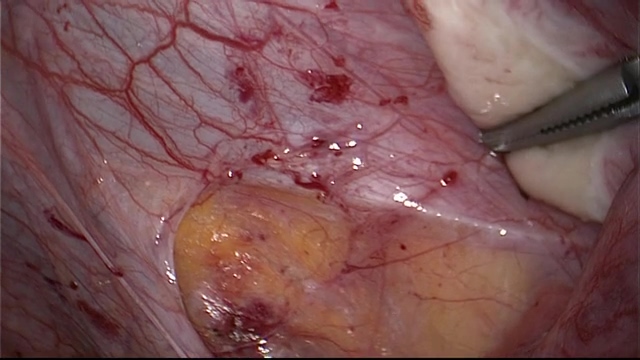}}
			\caption[]%
			{{\small}}    
			\label{fig:per_bot_0}
		\end{subfigure}
		\hskip0.5em\relax
		\begin{subfigure}[b]{0.18\textwidth}   
			\centering 
			\cfbox{black}{\includegraphics[width=\textwidth]{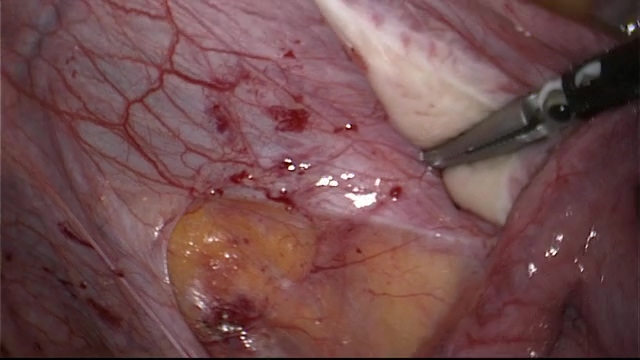}}
			\caption[]%
			{{\small}}    
			\label{fig:per_bot_1}
		\end{subfigure}
		\hskip0.5em\relax
		\begin{subfigure}[b]{0.18\textwidth}   
			\centering 
			\cfbox{black}{\includegraphics[width=\textwidth]{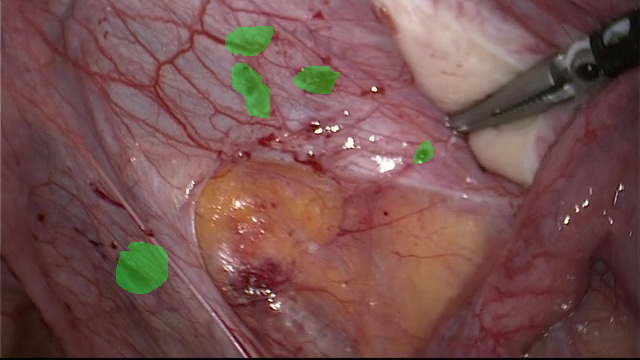}}
			\caption[]%
			{{\small}}    
			\label{fig:per_bot_2}
		\end{subfigure}
		\hskip0.5em\relax
		\begin{subfigure}[b]{0.18\textwidth}
			\centering
			\cfbox{black}{\includegraphics[width=\textwidth]{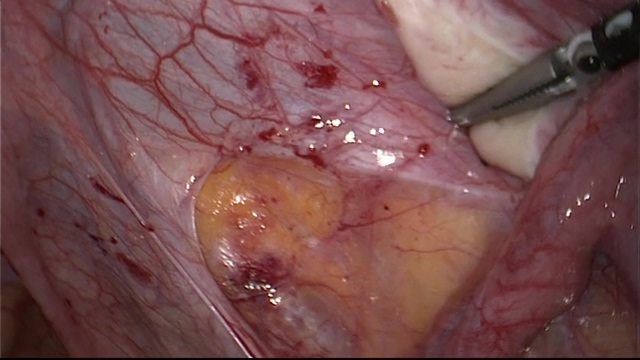}}
			\caption[]%
			{{\small}}    
			\label{fig:per_bot_3}
		\end{subfigure}
		\hskip0.5em\relax
		\begin{subfigure}[b]{0.18\textwidth}   
			\centering 
			\cfbox{black}{\includegraphics[width=\textwidth]{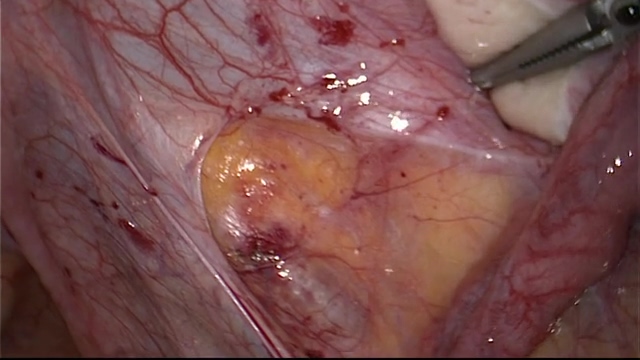}}
			\caption[]%
			{{\small}}    
			\label{fig:per_bot_4}
		\end{subfigure}
	\caption[]
	{\small Peritoneum: differing example images (\ref{fig:per_top_0} - \ref{fig:per_top_4}) with corresponding annotations (\ref{fig:per_mid_0} - \ref{fig:per_mid_4}) and video sequence example including keyframe annotations (\ref{fig:per_bot_0} - \ref{fig:per_bot_4}).} 
	\label{fig:class_per}  
	\vspace{-4mm}
\end{figure}  

\subsubsection{Peritoneum}
The peritoneum is a serous membrane lining the abdominal cavity (parietal) as well as its contained upper organs (visceral). Endometrial tissue annotated in GLENDA is associated with the pelvic cavity, which is enclosed by the parietal peritoneum. Figure~\ref{fig:class_per} shows various peritoneum dataset examples together with their annotations\footnote{Note that due to the possibility of annotating several categories per image, e.g. GLENDA includes region-based annotations of up to three classes per image, for simplicity only frames with exactly one associated class have been chosen as examples.} (binary images) and selected frames of a video sequence with its annotated keyframe (green overlay). Since the peritoneum covers a very large area and as well surrounds organs that are part of other GLENDA classes, corresponding images are often very different to one another and can contain non-relevant other areas that may even contain additional endometrial tissue. Another implication of the membrane's proportionally large size is its consequently very frequent visibility in many of the dataset's images, be it pathologoical or non pathological. Visually the peritoneum occurs in a mixture of red, yellow and white colors.

\begin{figure}[!htb]
	\begin{subfigure}[b]{0.18\textwidth}
		\centering
		\cfbox{black}{\includegraphics[width=\textwidth]{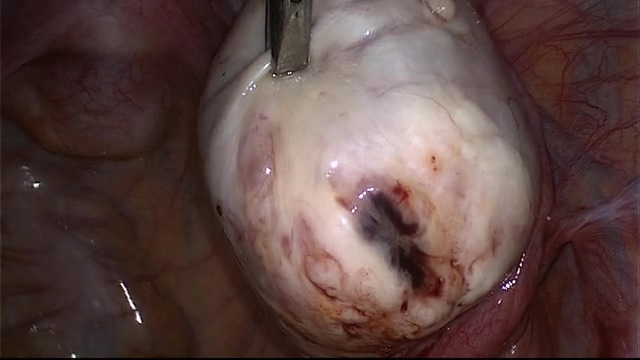}}
		\caption[]%
		{{\small}}    
		\label{fig:ovary_top_0}
	\end{subfigure}
	\hskip0.5em\relax
	\begin{subfigure}[b]{0.18\textwidth}   
		\centering 
		\cfbox{black}{\includegraphics[width=\textwidth]{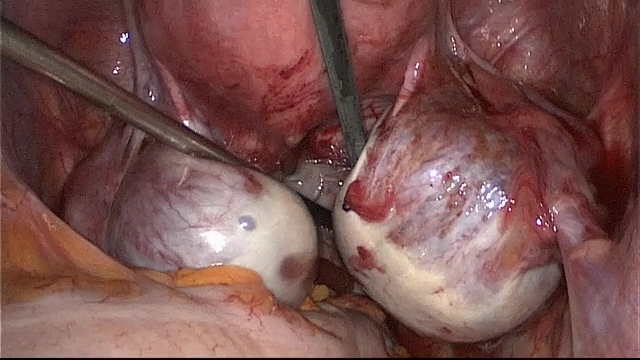}}
		\caption[]%
		{{\small}}    
		\label{fig:ovary_top_1}
	\end{subfigure}
	\hskip0.5em\relax
	\begin{subfigure}[b]{0.18\textwidth}   
		\centering 
		\cfbox{black}{\includegraphics[width=\textwidth]{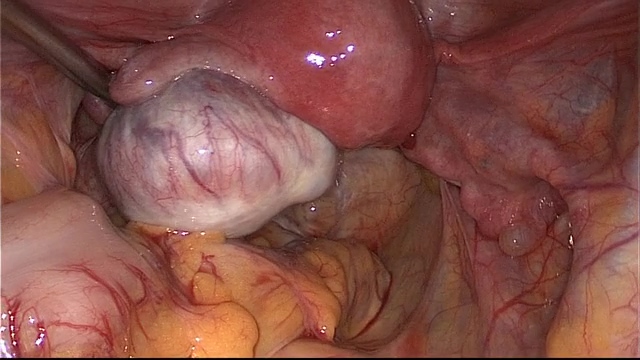}}
		\caption[]%
		{{\small}}    
		\label{fig:ovary_top_2}
	\end{subfigure}
	\hskip0.5em\relax
	\begin{subfigure}[b]{0.18\textwidth}
		\centering
		\cfbox{black}{\includegraphics[width=\textwidth]{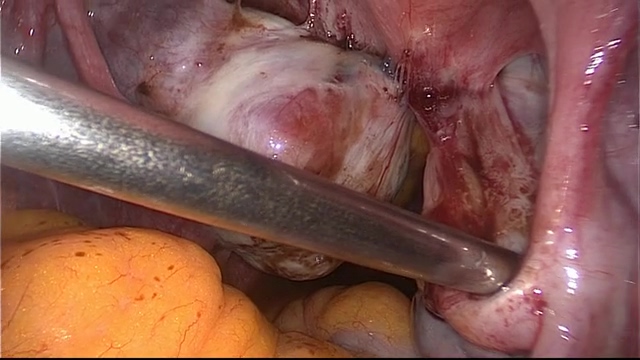}}
		\caption[]%
		{{\small}}    
		\label{fig:ovary_top_3}
	\end{subfigure}
	\hskip0.5em\relax
	\begin{subfigure}[b]{0.18\textwidth}   
		\centering 
		\cfbox{black}{\includegraphics[width=\textwidth]{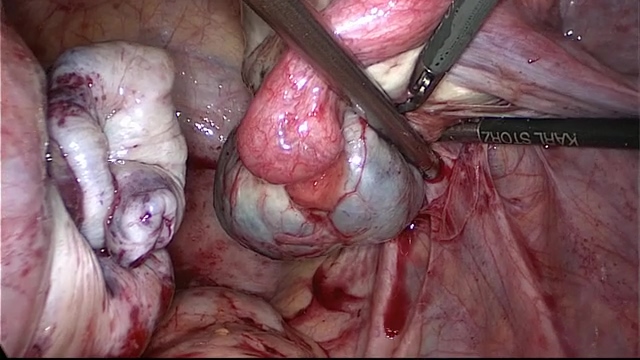}}
		\caption[]%
		{{\small}}    
		\label{fig:ovary_top_4}
	\end{subfigure}
	\vskip0.5em\relax
	\begin{subfigure}[b]{0.18\textwidth}
		\centering
		\cfbox{black}{\includegraphics[width=\textwidth]{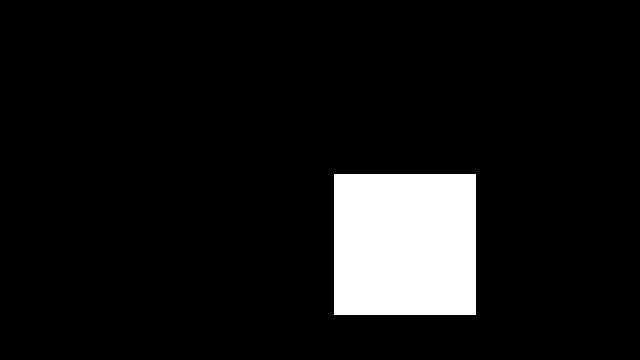}}
		\caption[]%
		{{\small}}    
		\label{fig:ovary_mid_0}
	\end{subfigure}
	\hskip0.5em\relax
	\begin{subfigure}[b]{0.18\textwidth}   
		\centering 
		\cfbox{black}{\includegraphics[width=\textwidth]{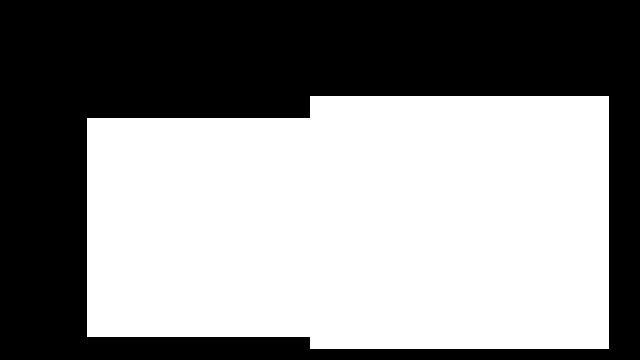}}
		\caption[]%
		{{\small}}    
		\label{fig:ovary_mid_1}
	\end{subfigure}
	\hskip0.5em\relax
	\begin{subfigure}[b]{0.18\textwidth}   
		\centering 
		\cfbox{black}{\includegraphics[width=\textwidth]{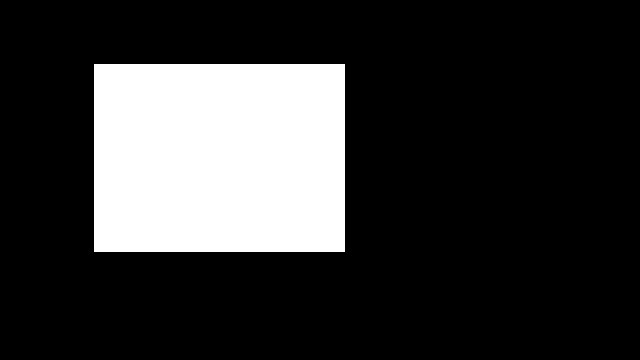}}
		\caption[]%
		{{\small}}    
		\label{fig:ovary_mid_2}
	\end{subfigure}
	\hskip0.5em\relax
	\begin{subfigure}[b]{0.18\textwidth}
		\centering
		\cfbox{black}{\includegraphics[width=\textwidth]{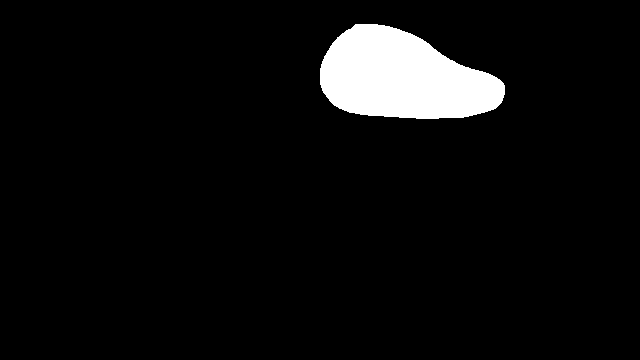}}
		\caption[]%
		{{\small}}    
		\label{fig:ovary_mid_3}
	\end{subfigure}
	\hskip0.5em\relax
	\begin{subfigure}[b]{0.18\textwidth}   
		\centering 
		\cfbox{black}{\includegraphics[width=\textwidth]{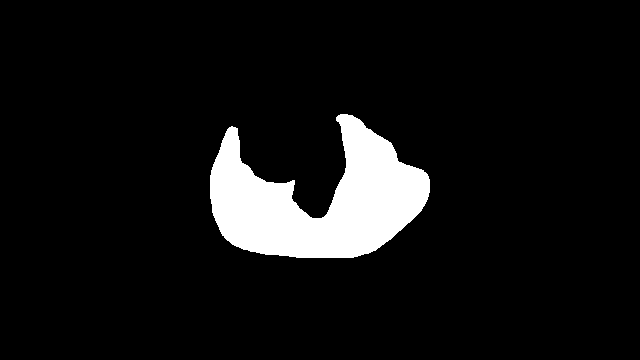}}
		\caption[]%
		{{\small}}    
		\label{fig:ovary_mid_4}
	\end{subfigure}
	\vskip0.5em\relax
		\begin{subfigure}[b]{0.18\textwidth}
			\centering
			\cfbox{black}{\includegraphics[width=\textwidth]{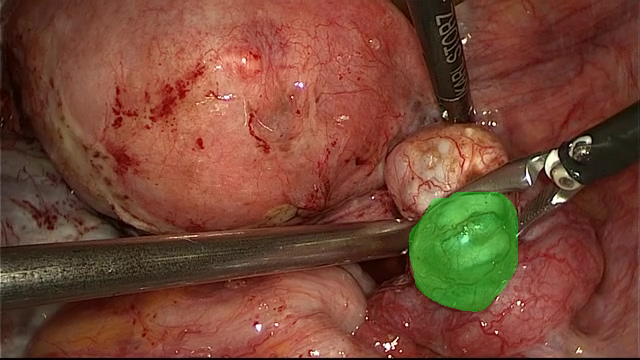}}
			\caption[]%
			{{\small}}    
			\label{fig:ovary_bot_0}
		\end{subfigure}
		\hskip0.5em\relax
		\begin{subfigure}[b]{0.18\textwidth}   
			\centering 
			\cfbox{black}{\includegraphics[width=\textwidth]{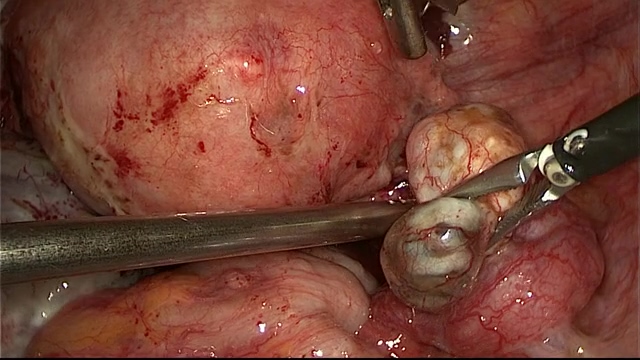}}
			\caption[]%
			{{\small}}    
			\label{fig:ovary_bot_1}
		\end{subfigure}
		\hskip0.5em\relax
		\begin{subfigure}[b]{0.18\textwidth}   
			\centering 
			\cfbox{black}{\includegraphics[width=\textwidth]{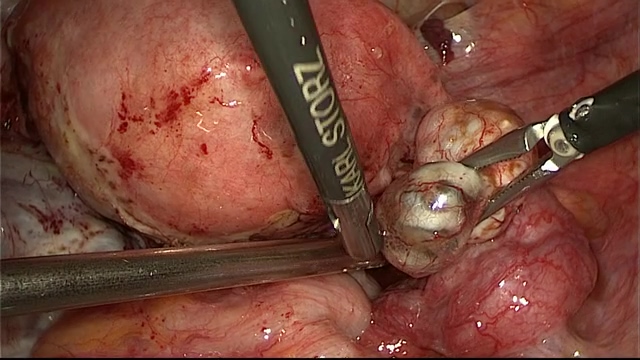}}
			\caption[]%
			{{\small}}    
			\label{fig:ovary_bot_2}
		\end{subfigure}
		\hskip0.5em\relax
		\begin{subfigure}[b]{0.18\textwidth}
			\centering
			\cfbox{black}{\includegraphics[width=\textwidth]{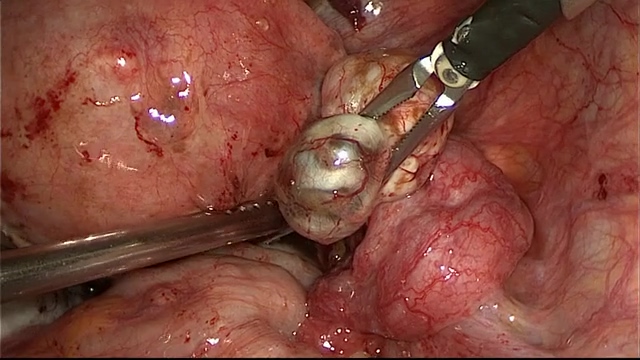}}
			\caption[]%
			{{\small}}    
			\label{fig:ovary_bot_3}
		\end{subfigure}
		\hskip0.5em\relax
		\begin{subfigure}[b]{0.18\textwidth}   
			\centering 
			\cfbox{black}{\includegraphics[width=\textwidth]{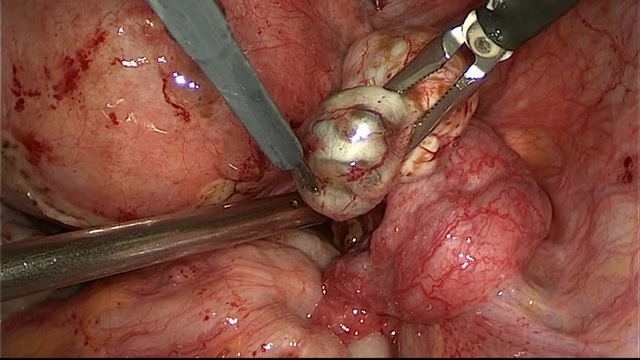}}
			\caption[]%
			{{\small}}    
			\label{fig:ovary_bot_4}
		\end{subfigure}
	\caption[]
	{\small Ovary: differing example images (\ref{fig:ovary_top_0} - \ref{fig:ovary_top_4}) with corresponding annotations (\ref{fig:ovary_mid_0} - \ref{fig:ovary_mid_4}) and video sequence example including keyframe annotations (\ref{fig:ovary_bot_0} - \ref{fig:ovary_bot_4}).} 
	\label{fig:class_ovary}  
	\vspace{-4mm}
\end{figure}  

\subsubsection{Ovary}
Apart from carrying several important functions like producing hormones, the main purpose of the two ovaries is to produce mature ova. Together with the peritoneum class, the ovaries are the most common locations for endometriosis (see Figure~\ref{fig:class_ovary} for dataset examples), which is the reason why both of them (together with the fallopian tubes) are the only lesion locations described by the rASRM score~\cite{canis1997revised}, specifically as well for diagnosing \textit{adhesions}, i.e. endometrial tissue connecting other tissue -- a class that, however, is not yet included in GLENDA. Visually ovaries are easily distinguishable from other organs even for laymen, since their oval-shaped outer capsule in non pathological state is colored in a shade of white contrasting them from the typical red-yellowish color spectrum of other anatomical structures.

\begin{figure}[!htb]
	\begin{subfigure}[b]{0.18\textwidth}
		\centering
		\cfbox{black}{\includegraphics[width=\textwidth]{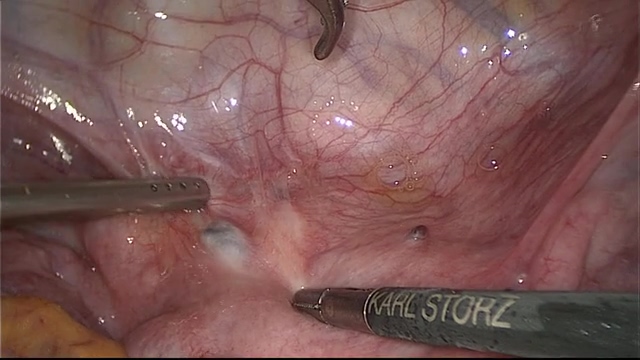}}
		\caption[]%
		{{\small}}    
		\label{fig:uterus_top_0}
	\end{subfigure}
	\hskip0.5em\relax
	\begin{subfigure}[b]{0.18\textwidth}   
		\centering 
		\cfbox{black}{\includegraphics[width=\textwidth]{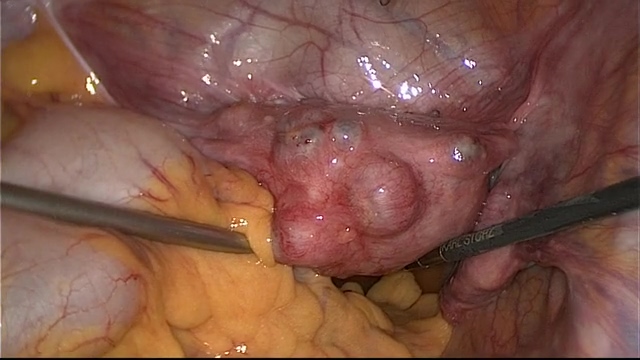}}
		\caption[]%
		{{\small}}    
		\label{fig:uterus_top_1}
	\end{subfigure}
	\hskip0.5em\relax
	\begin{subfigure}[b]{0.18\textwidth}   
		\centering 
		\cfbox{black}{\includegraphics[width=\textwidth]{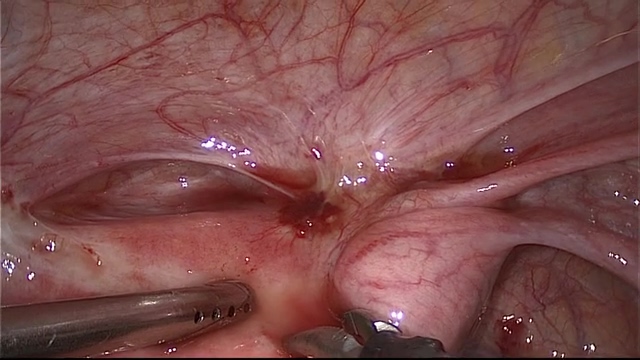}}
		\caption[]%
		{{\small}}    
		\label{fig:uterus_top_2}
	\end{subfigure}
	\hskip0.5em\relax
	\begin{subfigure}[b]{0.18\textwidth}
		\centering
		\cfbox{black}{\includegraphics[width=\textwidth]{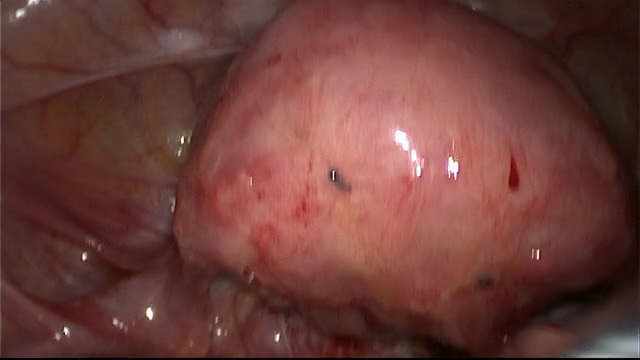}}
		\caption[]%
		{{\small}}    
		\label{fig:uterus_top_3}
	\end{subfigure}
	\hskip0.5em\relax
	\begin{subfigure}[b]{0.18\textwidth}   
		\centering 
		\cfbox{black}{\includegraphics[width=\textwidth]{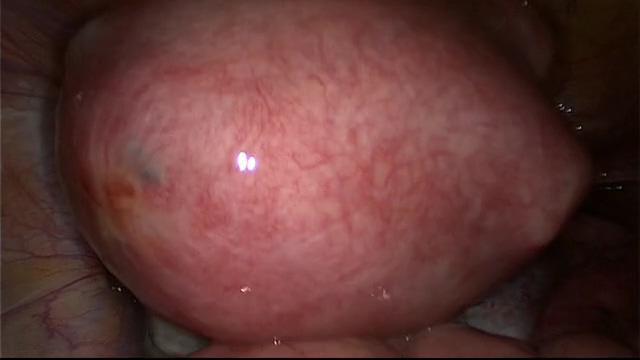}}
		\caption[]%
		{{\small}}    
		\label{fig:uterus_top_4}
	\end{subfigure}
	\vskip0.5em\relax
	\begin{subfigure}[b]{0.18\textwidth}
		\centering
		\cfbox{black}{\includegraphics[width=\textwidth]{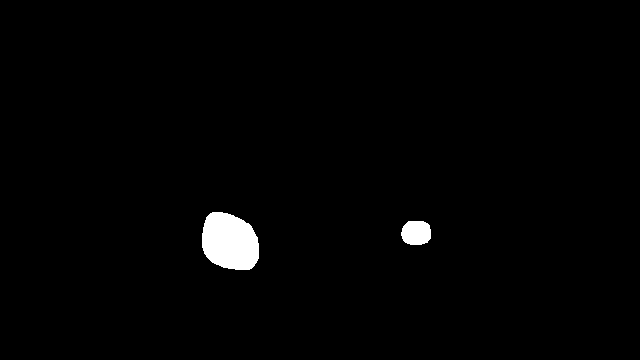}}
		\caption[]%
		{{\small}}    
		\label{fig:uterus_mid_0}
	\end{subfigure}
	\hskip0.5em\relax
	\begin{subfigure}[b]{0.18\textwidth}   
		\centering 
		\cfbox{black}{\includegraphics[width=\textwidth]{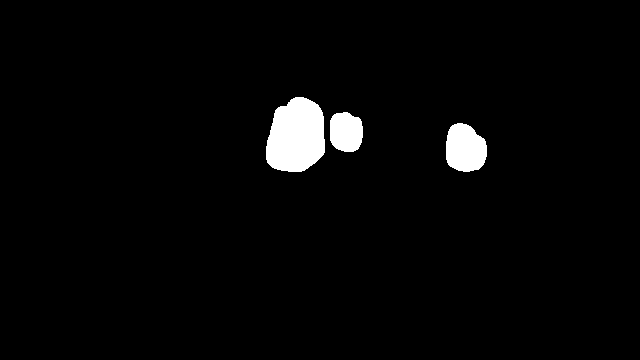}}
		\caption[]%
		{{\small}}    
		\label{fig:uterus_mid_1}
	\end{subfigure}
	\hskip0.5em\relax
	\begin{subfigure}[b]{0.18\textwidth}   
		\centering 
		\cfbox{black}{\includegraphics[width=\textwidth]{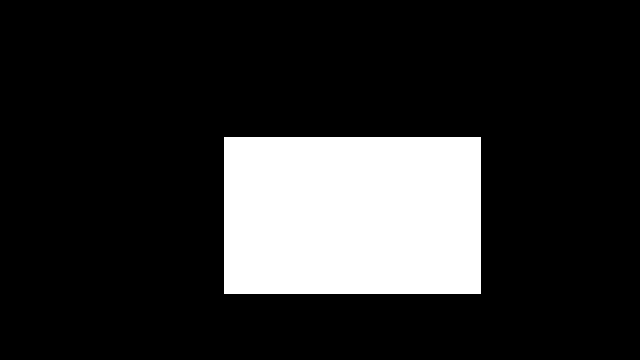}}
		\caption[]%
		{{\small}}    
		\label{fig:uterus_mid_2}
	\end{subfigure}
	\hskip0.5em\relax
	\begin{subfigure}[b]{0.18\textwidth}
		\centering
		\cfbox{black}{\includegraphics[width=\textwidth]{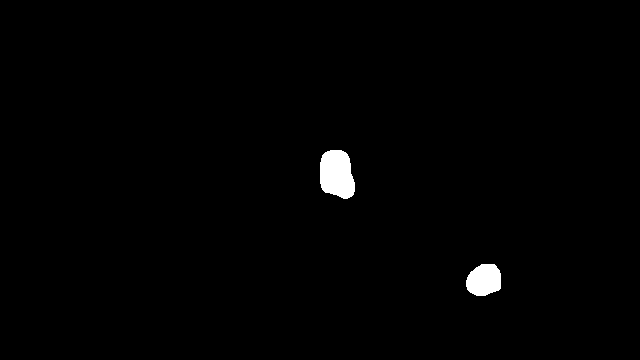}}
		\caption[]%
		{{\small}}    
		\label{fig:uterus_mid_3}
	\end{subfigure}
	\hskip0.5em\relax
	\begin{subfigure}[b]{0.18\textwidth}   
		\centering 
		\cfbox{black}{\includegraphics[width=\textwidth]{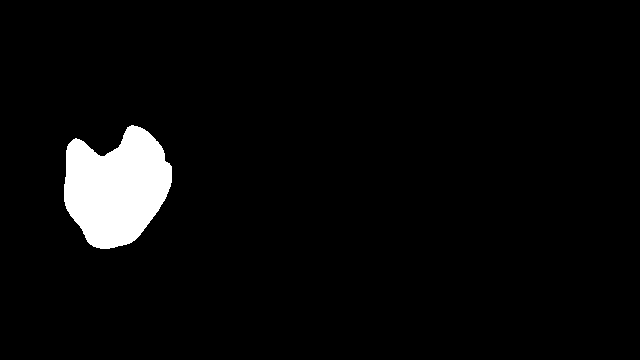}}
		\caption[]%
		{{\small}}    
		\label{fig:uterus_mid_4}
	\end{subfigure}
	\vskip0.5em\relax
		\begin{subfigure}[b]{0.18\textwidth}
			\centering
			\cfbox{black}{\includegraphics[width=\textwidth]{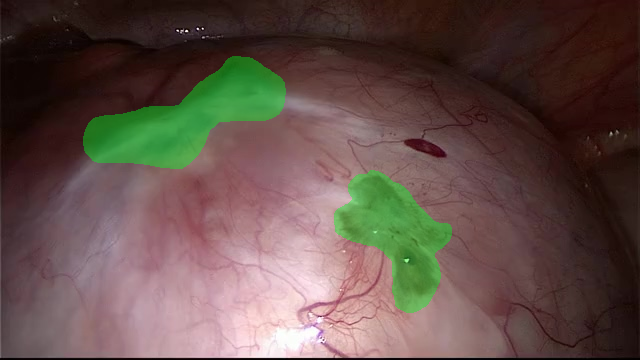}}
			\caption[]%
			{{\small}}    
			\label{fig:uterus_bot_0}
		\end{subfigure}
		\hskip0.5em\relax
		\begin{subfigure}[b]{0.18\textwidth}   
			\centering 
			\cfbox{black}{\includegraphics[width=\textwidth]{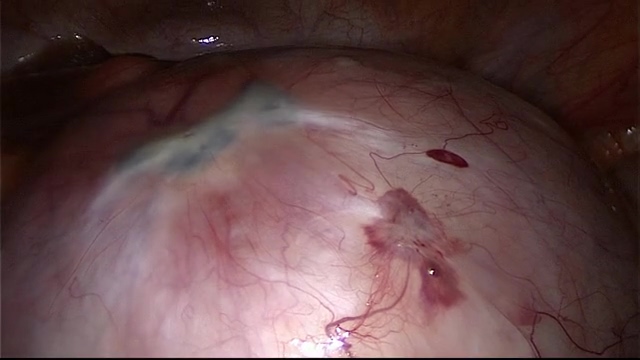}}
			\caption[]%
			{{\small}}    
			\label{fig:uterus_bot_1}
		\end{subfigure}
		\hskip0.5em\relax
		\begin{subfigure}[b]{0.18\textwidth}   
			\centering 
			\cfbox{black}{\includegraphics[width=\textwidth]{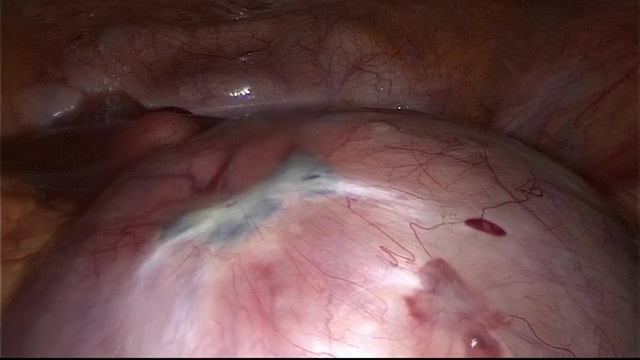}}
			\caption[]%
			{{\small}}    
			\label{fig:uterus_bot_2}
		\end{subfigure}
		\hskip0.5em\relax
		\begin{subfigure}[b]{0.18\textwidth}
			\centering
			\cfbox{black}{\includegraphics[width=\textwidth]{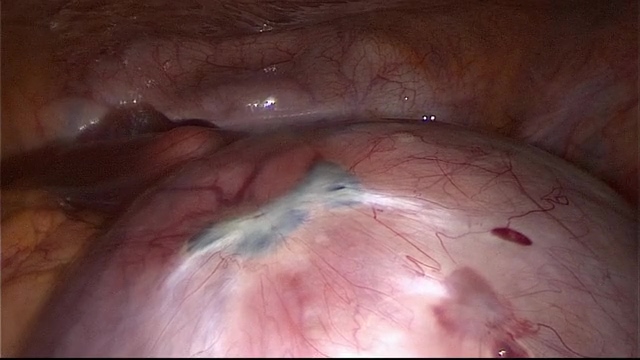}}
			\caption[]%
			{{\small}}    
			\label{fig:uterus_bot_3}
		\end{subfigure}
		\hskip0.5em\relax
		\begin{subfigure}[b]{0.18\textwidth}   
			\centering 
			\cfbox{black}{\includegraphics[width=\textwidth]{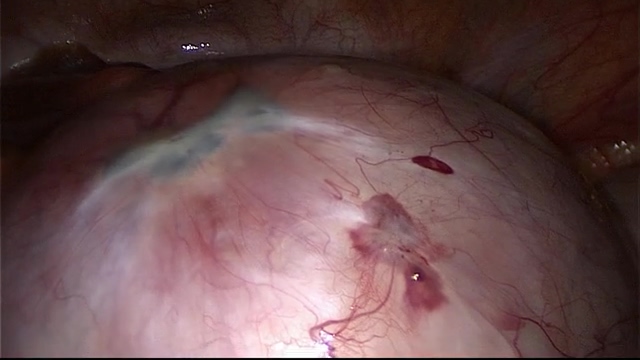}}
			\caption[]%
			{{\small}}    
			\label{fig:uterus_bot_4}
		\end{subfigure}
	\caption[]
	{\small Uterus: differing example images (\ref{fig:uterus_top_0} - \ref{fig:uterus_top_4}) with corresponding annotations (\ref{fig:uterus_mid_0} - \ref{fig:uterus_mid_4}) and video sequence example including keyframe annotations (\ref{fig:uterus_bot_0} - \ref{fig:uterus_bot_4}).} 
	\label{fig:class_uterus}
	\vspace{-4mm}
\end{figure}  

\subsubsection{Uterus}
The uterus is intended for bringing up a fetus from a fertilized ovum. Uteri can show different types of endometrial dislocation: endometrial tissue growing into the muscle wall of the uterus and thickening it is called \textit{adenomyosis} (adenomyosis may alter the shape and consistency of the uterus), while there is no special term for the case that the tissue is found on the uterine surface, which is covered by the visceral peritoneum. Similar to the previous classes, Figure~\ref{fig:class_uterus} depicts various examples for this category together with a sample sequence of an affected uterus. A non pathological uterus is pear-shaped, visually appears in shades of red and is located in-between the two ovaries and in connection to them via the fallopian tubes.

\begin{figure}[!htb]
	\begin{subfigure}[b]{0.18\textwidth}
		\centering
		\cfbox{black}{\includegraphics[width=\textwidth]{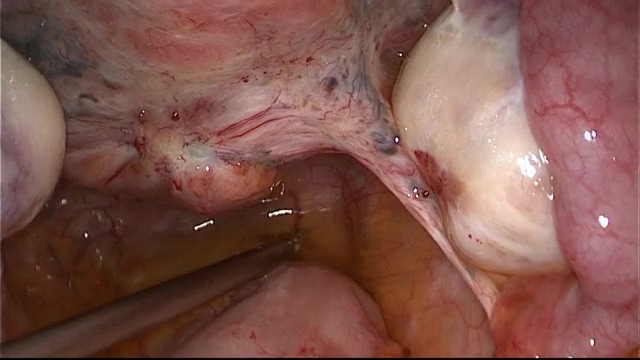}}
		\caption[]%
		{{\small}}    
		\label{fig:die_top_0}
	\end{subfigure}
	\hskip0.5em\relax
	\begin{subfigure}[b]{0.18\textwidth}   
		\centering 
		\cfbox{black}{\includegraphics[width=\textwidth]{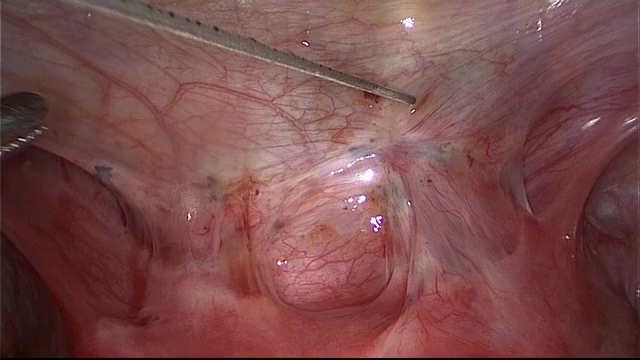}}
		\caption[]%
		{{\small}}    
		\label{fig:die_top_1}
	\end{subfigure}
	\hskip0.5em\relax
	\begin{subfigure}[b]{0.18\textwidth}   
		\centering 
		\cfbox{black}{\includegraphics[width=\textwidth]{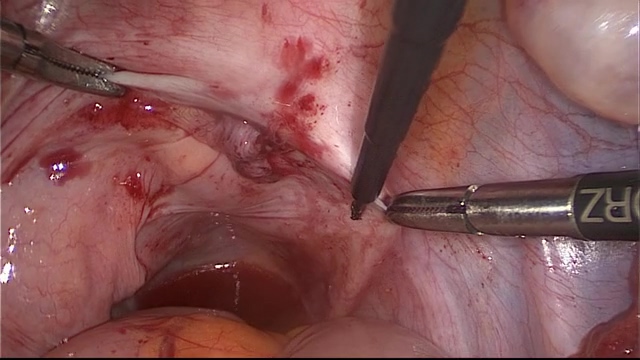}}
		\caption[]%
		{{\small}}    
		\label{fig:die_top_2}
	\end{subfigure}
	\hskip0.5em\relax
	\begin{subfigure}[b]{0.18\textwidth}
		\centering
		\cfbox{black}{\includegraphics[width=\textwidth]{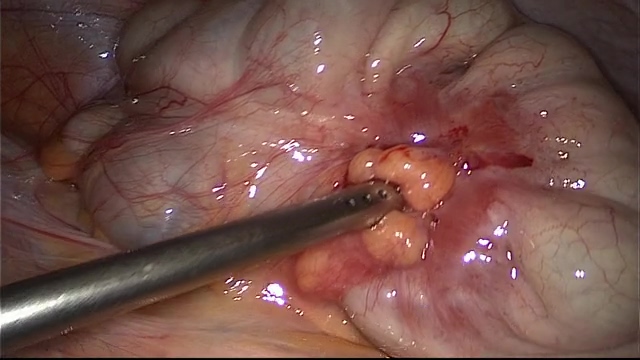}}
		\caption[]%
		{{\small}}    
		\label{fig:die_top_3}
	\end{subfigure}
	\hskip0.5em\relax
	\begin{subfigure}[b]{0.18\textwidth}   
		\centering 
		\cfbox{black}{\includegraphics[width=\textwidth]{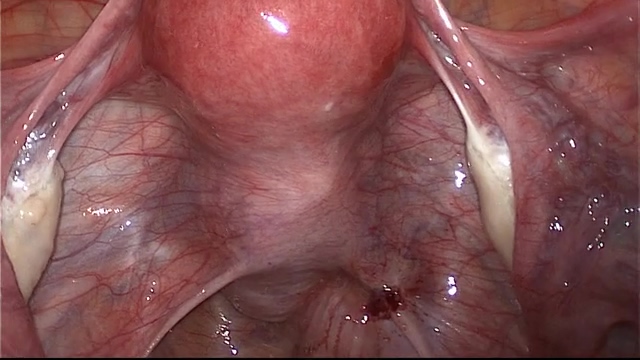}}
		\caption[]%
		{{\small}}    
		\label{fig:die_top_4}
	\end{subfigure}
	\vskip0.5em\relax
	\begin{subfigure}[b]{0.18\textwidth}
		\centering
		\cfbox{black}{\includegraphics[width=\textwidth]{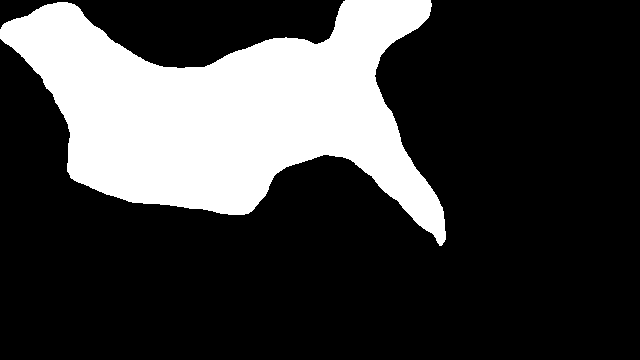}}
		\caption[]%
		{{\small}}    
		\label{fig:die_mid_0}
	\end{subfigure}
	\hskip0.5em\relax
	\begin{subfigure}[b]{0.18\textwidth}   
		\centering 
		\cfbox{black}{\includegraphics[width=\textwidth]{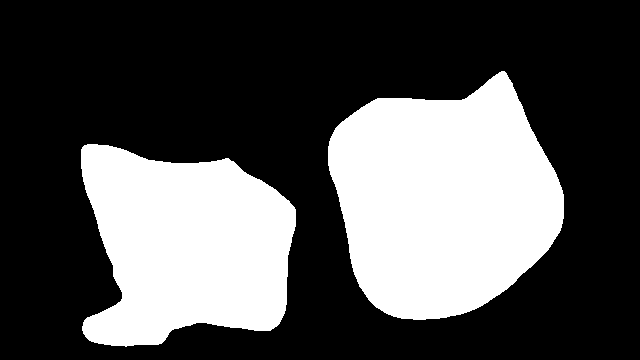}}
		\caption[]%
		{{\small}}    
		\label{fig:die_mid_1}
	\end{subfigure}
	\hskip0.5em\relax
	\begin{subfigure}[b]{0.18\textwidth}   
		\centering 
		\cfbox{black}{\includegraphics[width=\textwidth]{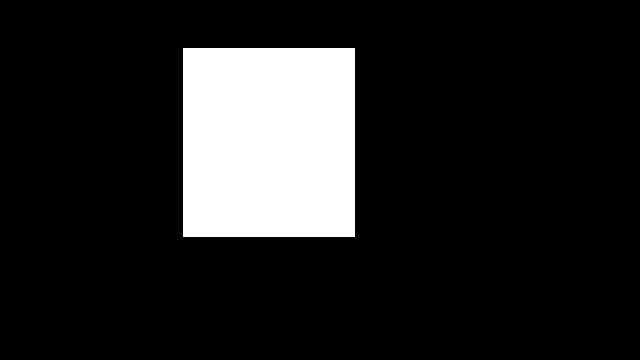}}
		\caption[]%
		{{\small}}    
		\label{fig:die_mid_2}
	\end{subfigure}
	\hskip0.5em\relax
	\begin{subfigure}[b]{0.18\textwidth}
		\centering
		\cfbox{black}{\includegraphics[width=\textwidth]{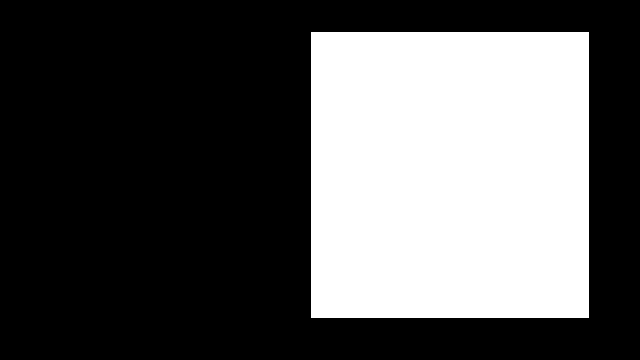}}
		\caption[]%
		{{\small}}    
		\label{fig:die_mid_3}
	\end{subfigure}
	\hskip0.5em\relax
	\begin{subfigure}[b]{0.18\textwidth}   
		\centering 
		\cfbox{black}{\includegraphics[width=\textwidth]{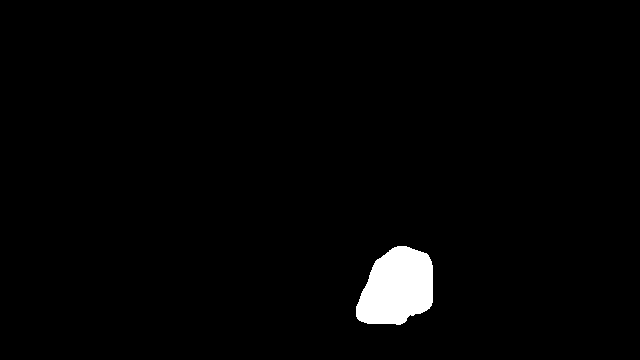}}
		\caption[]%
		{{\small}}    
		\label{fig:die_mid_4}
	\end{subfigure}
	\vskip0.5em\relax
		\begin{subfigure}[b]{0.18\textwidth}
			\centering
			\cfbox{black}{\includegraphics[width=\textwidth]{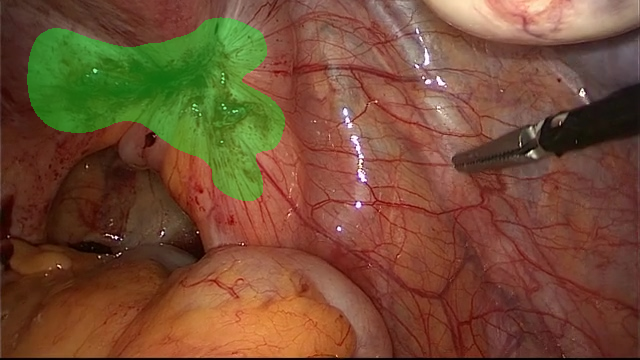}}
			\caption[]%
			{{\small}}    
			\label{fig:die_bot_0}
		\end{subfigure}
		\hskip0.5em\relax
		\begin{subfigure}[b]{0.18\textwidth}   
			\centering 
			\cfbox{black}{\includegraphics[width=\textwidth]{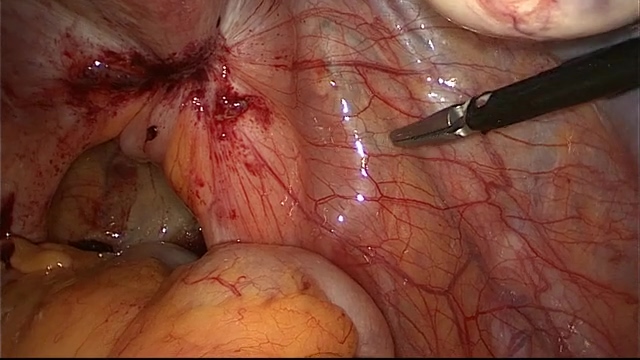}}
			\caption[]%
			{{\small}}    
			\label{fig:die_bot_1}
		\end{subfigure}
		\hskip0.5em\relax
		\begin{subfigure}[b]{0.18\textwidth}   
			\centering 
			\cfbox{black}{\includegraphics[width=\textwidth]{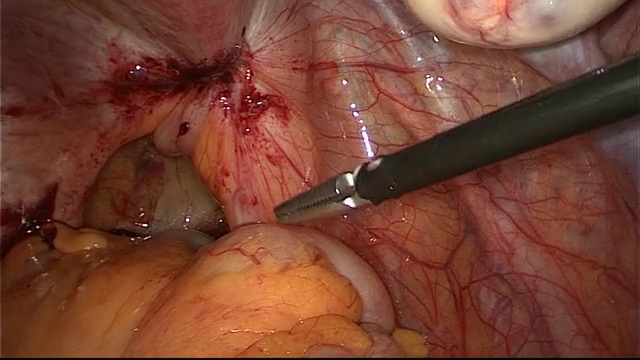}}
			\caption[]%
			{{\small}}    
			\label{fig:die_bot_2}
		\end{subfigure}
		\hskip0.5em\relax
		\begin{subfigure}[b]{0.18\textwidth}
			\centering
			\cfbox{black}{\includegraphics[width=\textwidth]{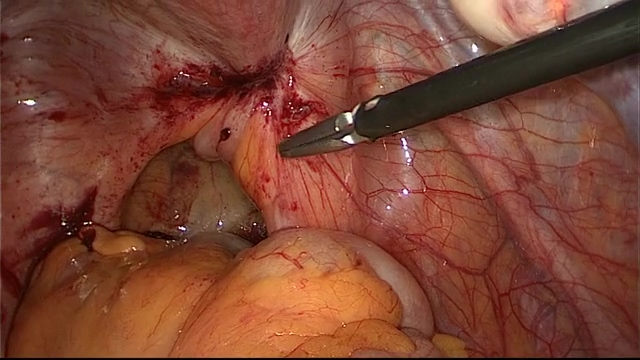}}
			\caption[]%
			{{\small}}    
			\label{fig:die_bot_3}
		\end{subfigure}
		\hskip0.5em\relax
		\begin{subfigure}[b]{0.18\textwidth}   
			\centering 
			\cfbox{black}{\includegraphics[width=\textwidth]{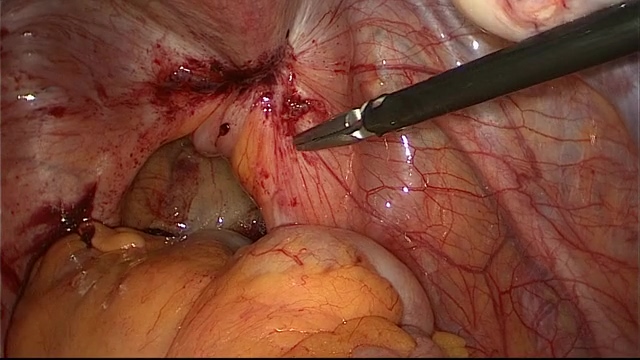}}
			\caption[]%
			{{\small}}    
			\label{fig:die_bot_4}
		\end{subfigure}
	\caption[]
	{\small Deep Infiltrating Endometriosis (DIE): differing example images (\ref{fig:die_top_0} - \ref{fig:die_top_4}) with corresponding annotations (\ref{fig:die_mid_0} - \ref{fig:die_mid_4}) and video sequence example including keyframe annotations (\ref{fig:die_bot_0} - \ref{fig:die_bot_4}).} \label{fig:class_die}
	\vspace{-4mm}
\end{figure}

\subsubsection{DIE}
Non-shallow endometriosis that is found on specific locations such as the rectum, the rectovaginal space or uterine ligaments is described as Deep Infiltrating Endometriosis (DIE) and is usually rated using the Enzian classification system~\cite{Keckstein2017} in addition to the rASRM score~\cite{canis1997revised}. Figure~\ref{fig:class_die} shows several examples of this class including lesions in the pelvic wall and uterine ligaments. Since all of the previously described anatomical structures can be affected by DIE rendering example pictures very similar to the other classes, it is a challenging task to classify DIE correctly. Additionally, as this type of endometriosis describes a large variety of lesion locations, no distinct visual appearance can be attributed to this type of class, other than highlighting that typically the color spectrum in recorded laparascopic videos lacks green tones.

\begin{figure}[!htb]
	\begin{subfigure}[b]{0.18\textwidth}
		\centering
		\cfbox{black}{\includegraphics[width=\textwidth]{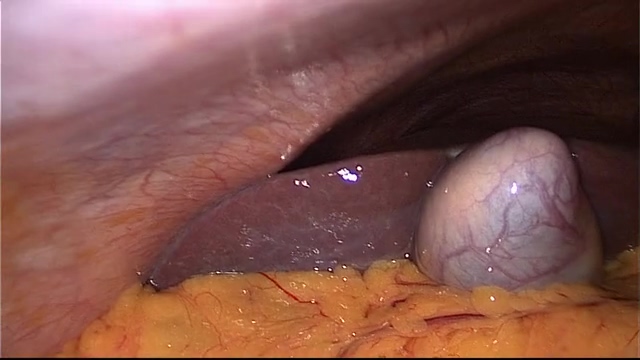}}
		\caption[]%
		{{\small}}    
		\label{fig:no_patho_top_0}
	\end{subfigure}
	\hskip0.5em\relax
	\begin{subfigure}[b]{0.18\textwidth}   
		\centering 
		\cfbox{black}{\includegraphics[width=\textwidth]{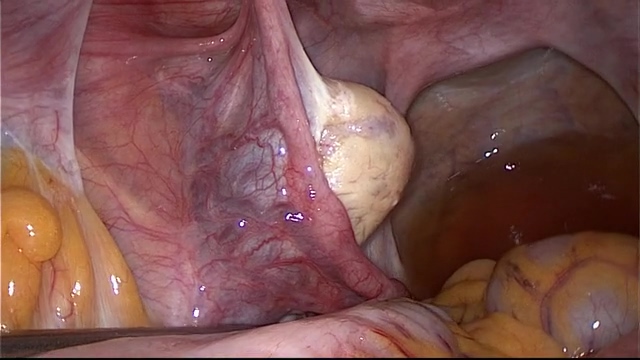}}
		\caption[]%
		{{\small}}    
		\label{fig:no_patho_top_1}
	\end{subfigure}
	\hskip0.5em\relax
	\begin{subfigure}[b]{0.18\textwidth}   
		\centering 
		\cfbox{black}{\includegraphics[width=\textwidth]{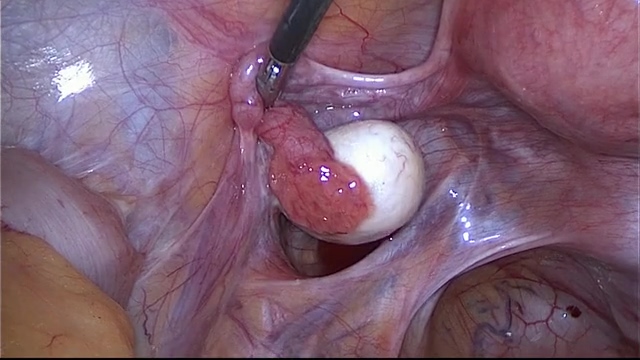}}
		\caption[]%
		{{\small}}    
		\label{fig:no_patho_top_2}
	\end{subfigure}
	\hskip0.5em\relax
	\begin{subfigure}[b]{0.18\textwidth}
		\centering
		\cfbox{black}{\includegraphics[width=\textwidth]{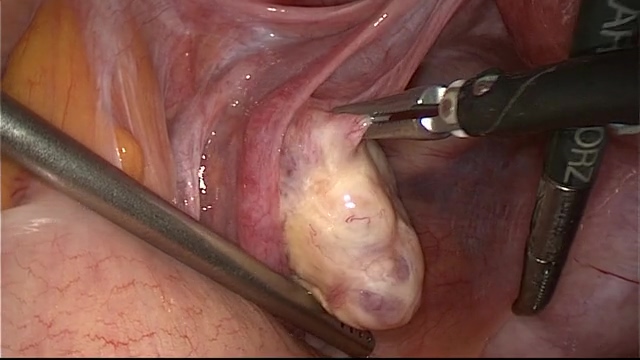}}
		\caption[]%
		{{\small}}    
		\label{fig:no_patho_top_3}
	\end{subfigure}
	\hskip0.5em\relax
	\begin{subfigure}[b]{0.18\textwidth}   
		\centering 
		\cfbox{black}{\includegraphics[width=\textwidth]{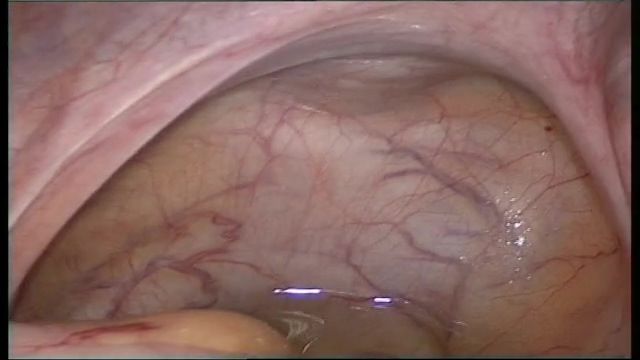}}
		\caption[]%
		{{\small}}    
		\label{fig:no_patho_top_4}
	\end{subfigure}
	\vskip0.5em\relax
	\begin{subfigure}[b]{0.18\textwidth}
		\centering
		\cfbox{black}{\includegraphics[width=\textwidth]{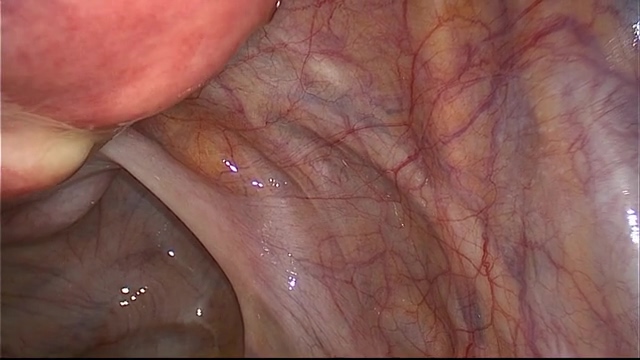}}
		\caption[]%
		{{\small}}    
		\label{fig:no_patho_mid_0}
	\end{subfigure}
	\hskip0.5em\relax
	\begin{subfigure}[b]{0.18\textwidth}   
		\centering 
		\cfbox{black}{\includegraphics[width=\textwidth]{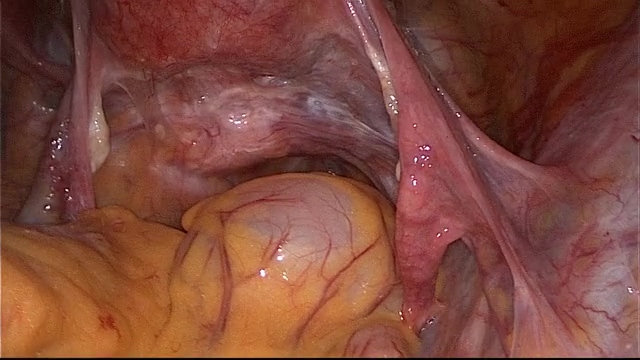}}
		\caption[]%
		{{\small}}    
		\label{fig:no_patho_mid_1}
	\end{subfigure}
	\hskip0.5em\relax
	\begin{subfigure}[b]{0.18\textwidth}   
		\centering 
		\cfbox{black}{\includegraphics[width=\textwidth]{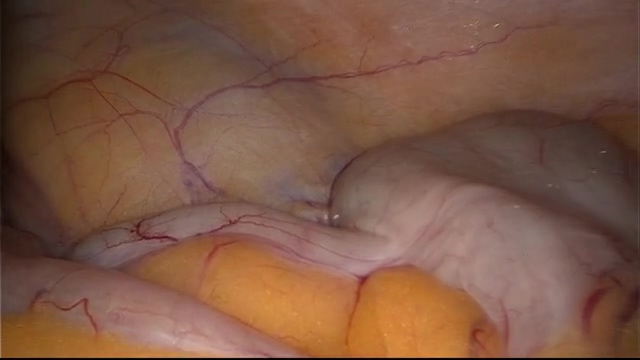}}
		\caption[]%
		{{\small}}    
		\label{fig:no_patho_mid_2}
	\end{subfigure}
	\hskip0.5em\relax
	\begin{subfigure}[b]{0.18\textwidth}
		\centering
		\cfbox{black}{\includegraphics[width=\textwidth]{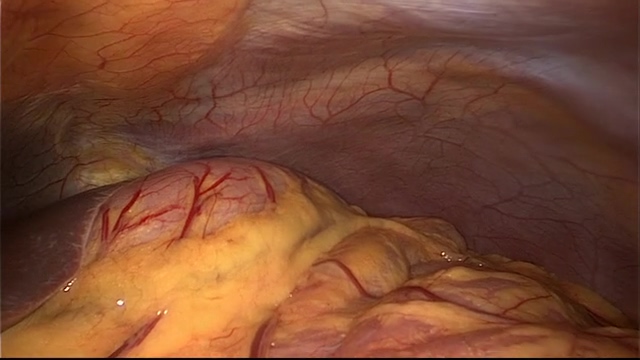}}
		\caption[]%
		{{\small}}    
		\label{fig:no_patho_mid_3}
	\end{subfigure}
	\hskip0.5em\relax
	\begin{subfigure}[b]{0.18\textwidth}   
		\centering 
		\cfbox{black}{\includegraphics[width=\textwidth]{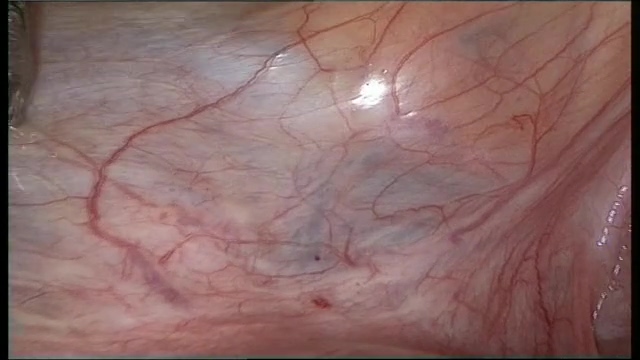}}
		\caption[]%
		{{\small}}    
		\label{fig:no_patho_mid_4}
	\end{subfigure}
	\vskip0.5em\relax
		\begin{subfigure}[b]{0.18\textwidth}
			\centering
			\cfbox{black}{\includegraphics[width=\textwidth]{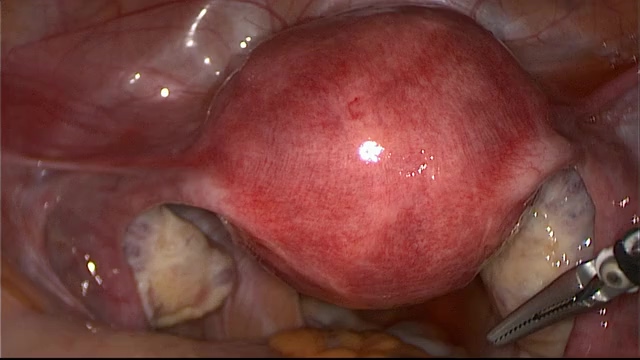}}
			\caption[]%
			{{\small}}    
			\label{fig:no_patho_bot_0}
		\end{subfigure}
		\hskip0.5em\relax
		\begin{subfigure}[b]{0.18\textwidth}   
			\centering 
			\cfbox{black}{\includegraphics[width=\textwidth]{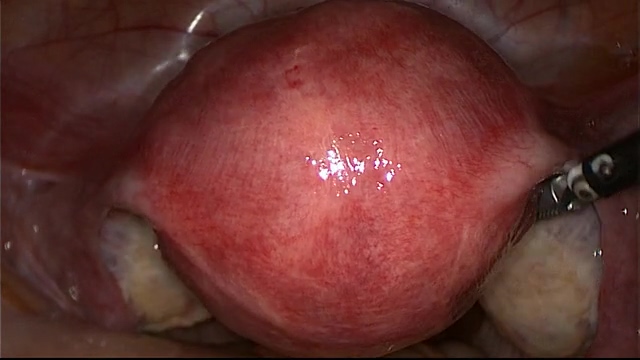}}
			\caption[]%
			{{\small}}    
			\label{fig:no_patho_bot_1}
		\end{subfigure}
		\hskip0.5em\relax
		\begin{subfigure}[b]{0.18\textwidth}   
			\centering 
			\cfbox{black}{\includegraphics[width=\textwidth]{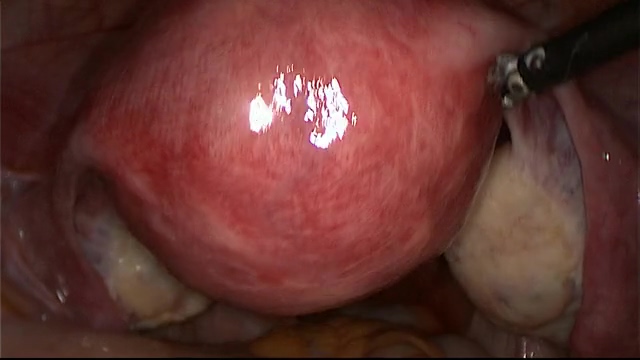}}
			\caption[]%
			{{\small}}    
			\label{fig:no_patho_bot_2}
		\end{subfigure}
		\hskip0.5em\relax
		\begin{subfigure}[b]{0.18\textwidth}
			\centering
			\cfbox{black}{\includegraphics[width=\textwidth]{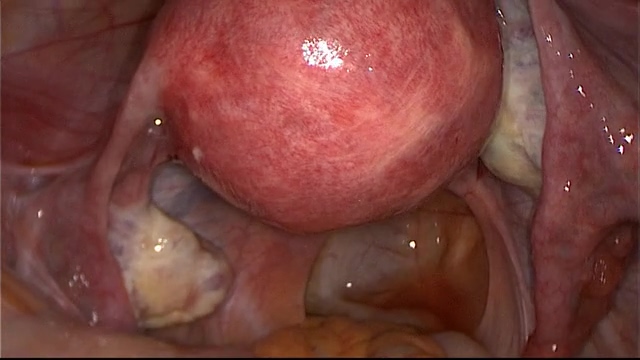}}
			\caption[]%
			{{\small}}    
			\label{fig:no_patho_bot_3}
		\end{subfigure}
		\hskip0.5em\relax
		\begin{subfigure}[b]{0.18\textwidth}   
			\centering 
			\cfbox{black}{\includegraphics[width=\textwidth]{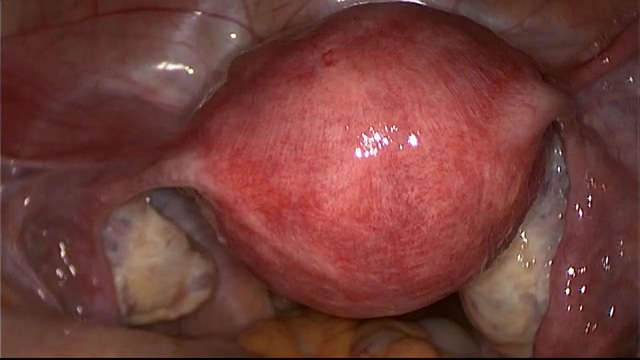}}
			\caption[]%
			{{\small}}    
			\label{fig:no_patho_bot_4}
		\end{subfigure}
	\caption[]
	{\small No Pathology: differing example images (\ref{fig:no_patho_top_0} - \ref{fig:no_patho_mid_4}) and video sequence example (\ref{fig:no_patho_bot_0} - \ref{fig:no_patho_bot_4}).} 
	\label{fig:class_no_patho}
	\vspace{-4mm}
\end{figure}  

\subsubsection{No Pathology}
Video sequences containing no visible pathology in relation to endometriosis are included in the dataset, providing counter examples to above categories. Since this class does not contain any region-based annotations, in addition to a sequence showing a non-pathological uterus, Figure~\ref{fig:class_no_patho} particularly includes examples of several anatomical structures from above pathological classes (e.g. peritoneum and ovaries). Again it is not possible to make any assumptions about the color and shape of objects within this class, since it includes images covering most areas of the pelvic region.

\subsection{Structure}
\label{sec:ds_structure}

\begin{figure}[!htb]
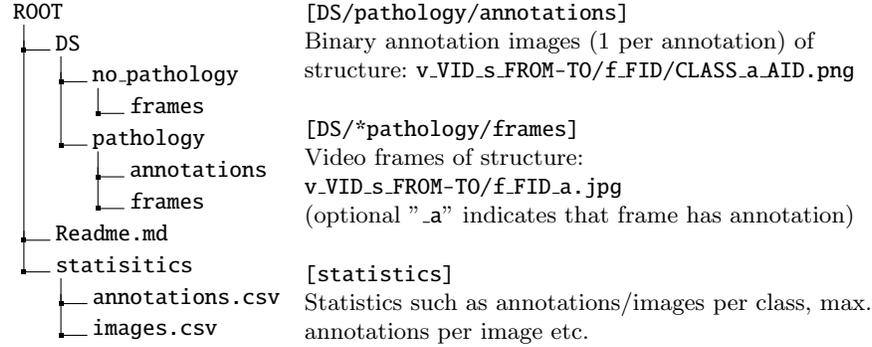

\begin{minipage}{.33\textwidth}
\dirtree{%
.1 ROOT.
.2 DS.
.3 no\_pathology.
.4 frames.
.3 pathology.
.4 annotations.
.4 frames.
.2 Readme.md.
.2 statisitics.
.3 annotations.csv.
.3 images.csv.
}
\end{minipage}
\begin{minipage}{0.65\textwidth}
    \begin{tabular}{>{\raggedright}p{\linewidth}}
    \hiderowcolors
    \leavevmode \\
    \texttt{[DS/pathology/annotations]}\\
    Binary annotation images (1 per annotation) of structure: \texttt{v\_VID\_s\_FROM-TO/f\_FID/CLASS\_a\_AID.png}\\
    \leavevmode \\
    \texttt{[DS/*pathology/frames]}\\
    Video frames of structure: \texttt{v\_VID\_s\_FROM-TO/f\_FID\_a.jpg}\\(optional "\texttt{\_a}" indicates that frame has annotation)\\
    \leavevmode \\
    \texttt{[statistics]}\\
    Statistics such as annotations/images per class, max. annotations per image etc.\\
    \end{tabular}
\end{minipage}
\caption{GLENDA's directory structure (placeholders capitalized).}
\label{fig:glenda_dir_struct}
\end{figure}

Although originally extracted from videos, GLENDA is an image-based dataset, hence, contains a structured collection of images. Besides images of video frames, all annotations have been extracted separately and are also provided as images, albeit in binary format (as depicted in Figures~\ref{fig:class_per}-\ref{fig:class_die}, but with the restriction of one annotation per image). The dataset archive addionally includes a Readme file as well as some dataset statistics in comma separated value (CSV) tables. Its directory structure is listed and explained in Figure~\ref{fig:glenda_dir_struct}.

GLENDA's folder structure as well as file names reflect all relevant information for utilizing the dataset. In particular every contained video, frame and annotation possesses a unique ascending ID number and frames can be mapped to their annotations by partial path matching, which can be sped up by first removing all non-annotated frames, i.e. finding frames with an "\texttt{\_a}" suffix:

\begin{Verbatim}[commandchars=\\\{\}]
DS/pathology/frames/\textbf{v_2401_s_210-318/f_210}_a.jpg ->
    DS/pathology/annotations/\textbf{v_2401_s_210-318/f_210}/die_a_629.png
    DS/pathology/annotations/\textbf{v_2401_s_210-318/f_210}/die_a_630.png
    DS/pathology/annotations/\textbf{v_2401_s_210-318/f_210}/die_a_631.png
    ...
\end{Verbatim}

Region-based annotations, which, as mentioned above, are in binary format (black background, white annotation) need to appropriately be transformed into the required format, e.g. rectanglar bounding boxes, polygons etc. For conducting binary classification region-based annotations are disregarded, hence, images can simply be retrieved from their corresponding folders (\texttt{DS/pathology/frames} and \texttt{DS/no\_pathology/frames}).


\section{Limitations}
\label{sec:limitations}

When thoroughly examining GLENDA in its current version, several shortcomings can be identified. Although the total amount of images is approximately balanced between examples for pathology and no pathology (12K+ vs. 13K+), single pathology classes are uneven: e.g. peritoneum contains more annotations than all three other classes combined. Especially the sample count of the uterus category with a total of merely 8 annotated images with 14 annotations is rather low. Therefore, researchers may either choose to omit this class altogether or attempt to augment it via tracking the annotations over the class' 5 included video segments. Although this will yield a bigger sample count of 475 images, it is of course less diverse than creating new annotations.

Paying even more attention to data diversity, when using the full 25K+ image corpus with the goal of machine learning based classification, it is important to carefully consider data preparation: sequential video frames within small periods of time (e.g. a few seconds) are very similar to each other, which can have a grave detrimental impact on the training process. Thus, training, validation and test splits should be constructed from distinct video sequences rather than from the entire image pool, specifically for non pathological samples, which exclusively are comprised of merely 27 sequences. Although more effortful, this approach has the advantage that the data across the splits is always truly different, preventing the classifier from validating/testing on exactly what it has been trained on, yielding a perfect classification score. In order to increase the amount of dissimilar sequences, it may also be feasible to uniformly sample from the existing sequences with a large enough interval or apply a shot detection technique for finding boundaries between dissimilar content. 

Finally, not least due to the varied sample count for pathological classes, in order to ensure an optimal training split that is balanced similarly to the dataset, it should be proportional to the amount of examples per class and not the total number of images, i.e. every split should contain a certain percentage of examples from each class.

\section{Conclusion}
\label{sec:conclusion}

With the aim of inspiring research in gynecologic laparoscopy, we introduce the first Gynecologic Laparoscy ENdometriosis DAtaset (GLENDA), created in collaboration with medical experts in the domain of endometriosis treatment. GLENDA contains over 25K images, about half of which are pathological, i.e. showing endometriosis, and the other half non-pathological, i.e. containing no visible endometriosis. We thoroughly describe the data collection process, the dataset's properties and structure, while also discussing its limitations. We plan on continuously extending GLENDA, including the addition of other relevant categories and ultimately lesion severities. Furthermore, we are in the process of collecting specific "endometriosis suspicion" class annotations in all categories for capturing a common situation among endometriosis experts: at times it proves difficult even for specialists to classify the anomaly without further inspection, which may be due to several reasons, such as visible video artifacts or DIE regions with a small surface area. Although doubling the amount of classes, having such difficult examples may greatly improve the quality of endometriosis classifiers.

\section*{Acknowledgements}
\small{
This work was funded by the FWF Austrian Science Fund under grant P 32010-N38.
}

%
%
%
\bibliographystyle{splncs04}
\bibliography{thebib}

\end{document}